\documentclass[letterpaper, 10 pt, conference]{ieeeconf}  %

\IEEEoverridecommandlockouts                              %

\usepackage{adjustbox}
\usepackage{graphicx}
\usepackage{amsmath}
\usepackage{amssymb}
\usepackage{booktabs}
\usepackage[english]{babel}

\usepackage{bm}
\usepackage{multirow}
\usepackage{subcaption}
\usepackage{algorithm}
\usepackage{algpseudocode}
\usepackage{float}
\usepackage[dvipsnames]{xcolor}
\usepackage{lipsum}
\usepackage{bbm}
\usepackage[none]{hyphenat}
\usepackage{array}
\usepackage{arydshln}  %

\usepackage{tikz}
\usetikzlibrary{positioning, fit}
\usepackage{cuted}
\usepackage[font=footnotesize]{caption}

\definecolor{cvprblue}{rgb}{0.21,0.49,0.74}
\definecolor{waymogreen}{HTML}{00E89D} %
\definecolor{waymoblue}{HTML}{0077FF} %
\definecolor{waymopurple}{HTML}{9150C8} %
\definecolor{waymoamber}{HTML}{FFCD55} %

\usepackage[pagebackref,breaklinks]{hyperref}

\newcommand{\ours}{Drive\&Gen}

\title{\scalebox{.92}[1.0]{\LARGE \bf
\ours{}: Co-Evaluating End-to-End Driving and Video Generation Models}}

\author{\scalebox{.96}[1.0]{Jiahao Wang$^{1}$, Zhenpei Yang$^{2}$, Yijing Bai$^{2}$, Yingwei Li$^{2}$, Yuliang Zou$^{2}$, Bo Sun$^{2}$, Abhijit Kundu$^{3}$, Jose Lezama$^{3}$}, \\\scalebox{.96}[1.0]{Luna Yue Huang$^{2}$, Zehao Zhu$^{2}$, Jyh-Jing Hwang$^{2}$, Dragomir Anguelov$^{2}$, Mingxing Tan$^{2}$, Chiyu Max Jiang$^{2}$}%
\thanks{*This work was supported by Waymo.}%
\thanks{$^{1}$Jiahao Wang is with Johns Hopkins University. Work done at Waymo. E-Mail: {\tt\small jiahaowg@gmail.com}}%
\thanks{$^{2}$Authors are with Waymo.}%
\thanks{$^{3}$Authors are with Google DeepMind.}%
}

\begin{document}

\maketitle

\hyphenpenalty=1000

\begin{strip}
\vspace{-8em}
\begin{center}
    \centering
    \begin{tikzpicture}[node distance=0.1cm]
    \newcommand{\imagesize}{0.135\textwidth} %
    \newcommand{\textgap}{0.25cm} %
    \newcommand{\imageshift}{-0.04*\imagesize}  %
    \newcommand{\imageshiftt}{-0.08*\imagesize}  %
    \tikzstyle{dotted}= [dash pattern=on 6\pgflinewidth off 1mm] 

    \node (start) {};
    \fboxsep=0pt
    \node[right=of start, z=100] (layout1) {\fbox{\includegraphics[trim={3px 3px 3px 3px},clip,width=\imagesize, height=\imagesize]{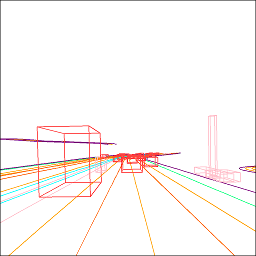}}};
    \node[right=of start, xshift=\imageshift, yshift=\imageshift] (layout2) {\fbox{\includegraphics[trim={3px 3px 3px 3px},clip,width=\imagesize, height=\imagesize]{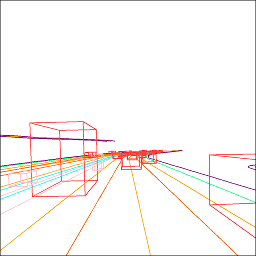}}};
    \node[right=of start, xshift=\imageshiftt, yshift=\imageshiftt] (layout3)
    {\fbox{\includegraphics[trim={3px 3px 3px 3px},clip,width=\imagesize, height=\imagesize]{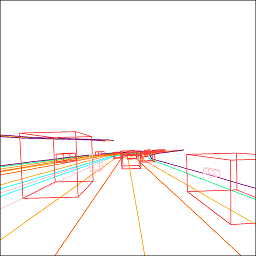}}};
    \node[right=-0.1cm of layout1] (gap1) {};
    \node[right=of gap1, z=100] (log1) {\fbox{\includegraphics[trim={3px 3px 3px 3px},clip,width=\imagesize, height=\imagesize]{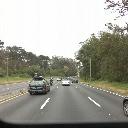}}};
    \node[right=of gap1, xshift=\imageshift, yshift=\imageshift] (log2) {\fbox{\includegraphics[trim={3px 3px 3px 3px},clip,width=\imagesize, height=\imagesize]{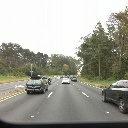}}};
    \node[right=of gap1, xshift=\imageshiftt, yshift=\imageshiftt] (log3) {\fbox{\includegraphics[trim={3px 3px 3px 3px},clip,width=\imagesize, height=\imagesize]{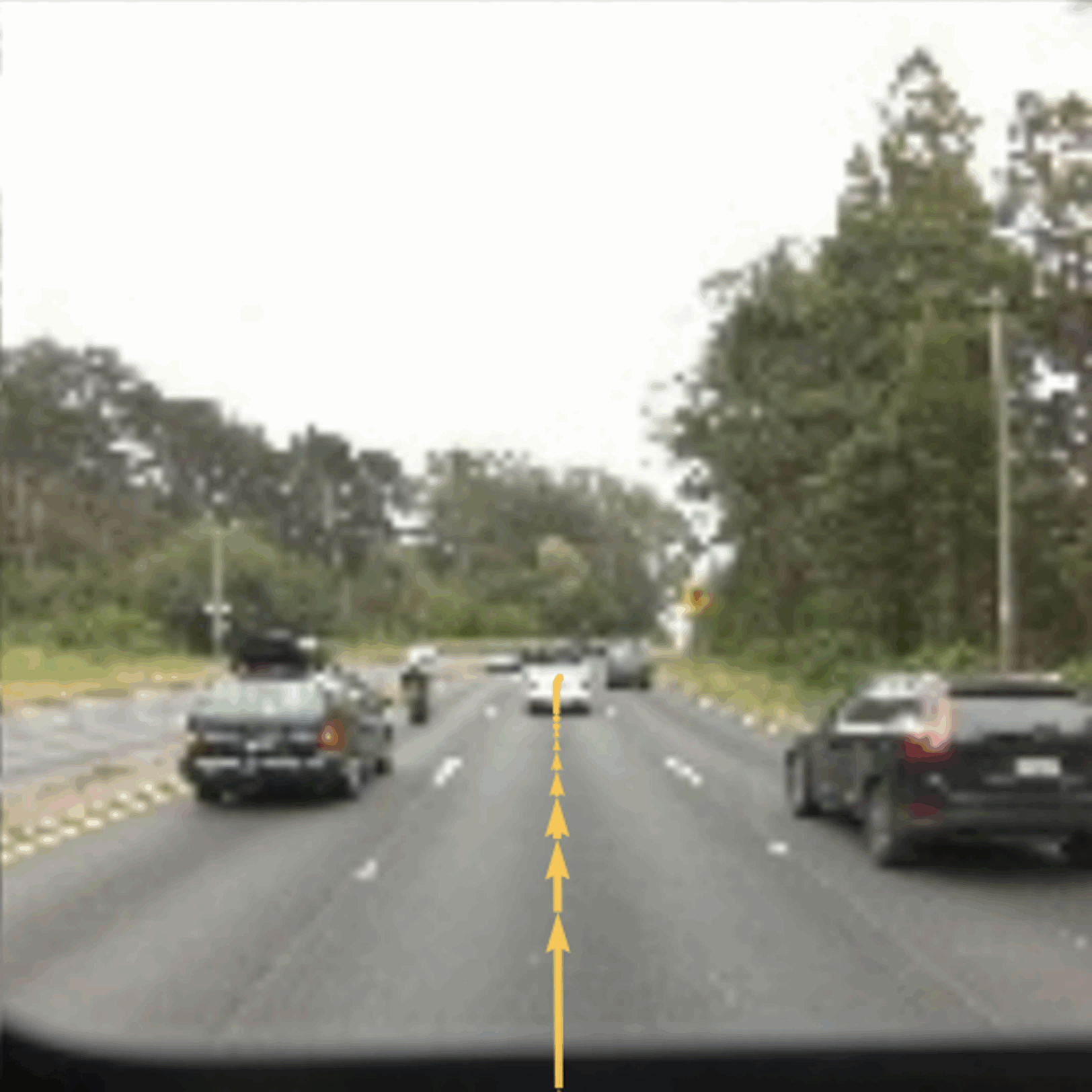}}};
    \node[right=of log1, z=100] (same1) {\fbox{\includegraphics[trim={3px 3px 3px 3px},clip,width=\imagesize, height=\imagesize]{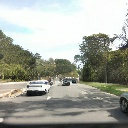}}};
    \node[right=of log1, xshift=\imageshift, yshift=\imageshift] (same2) {\fbox{\includegraphics[trim={3px 3px 3px 3px},clip,width=\imagesize, height=\imagesize]{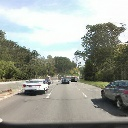}}};
    \node[right=of log1, xshift=\imageshiftt, yshift=\imageshiftt] (same3) {\fbox{\includegraphics[trim={3px 3px 3px 3px},clip,width=\imagesize, height=\imagesize]{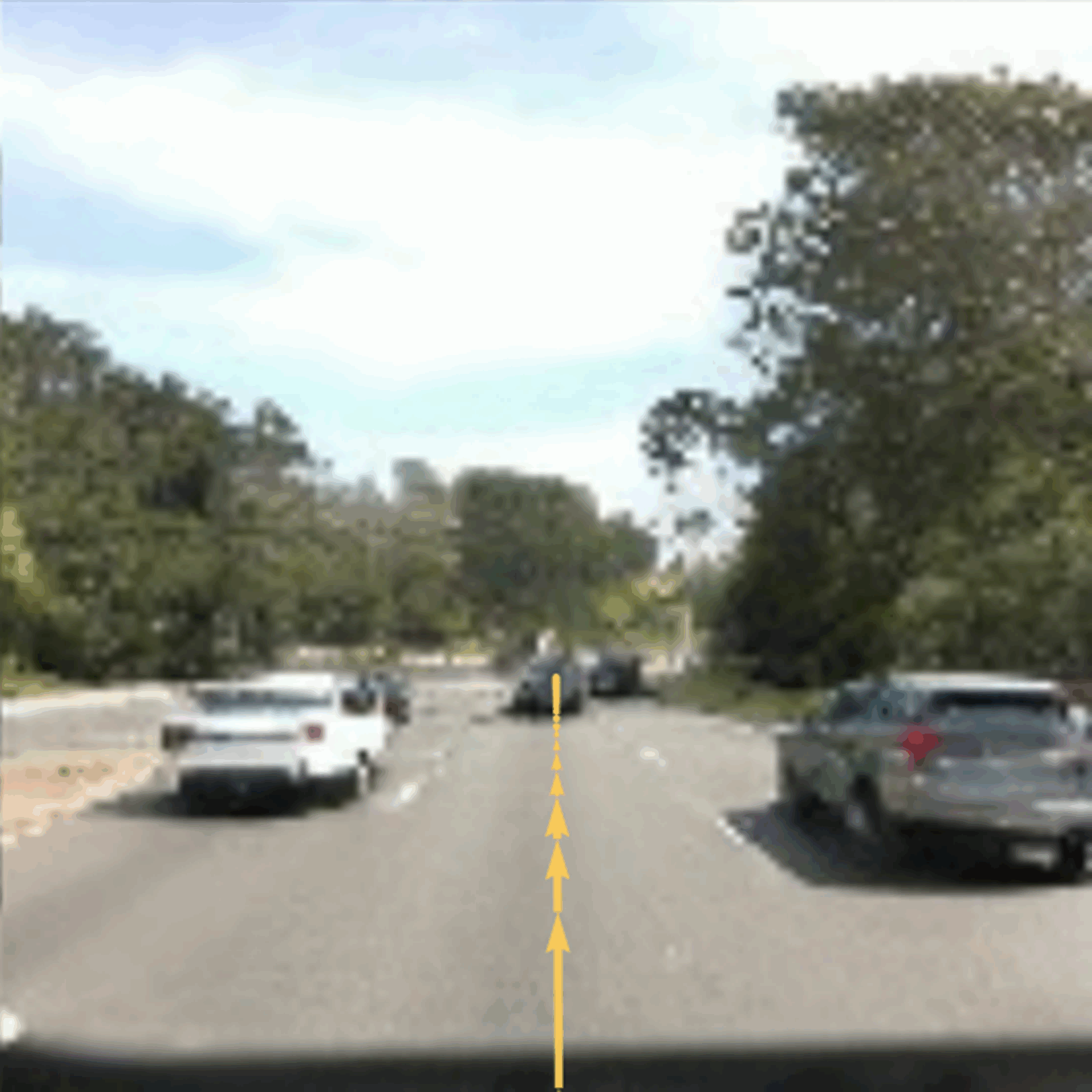}}};
    \node[right=of same1] (gap2) {};
    \node[right=of gap2, z=100] (weather1) {\fbox{\includegraphics[trim={3px 3px 3px 3px},clip,width=\imagesize, height=\imagesize]{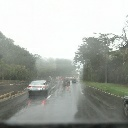}}};
    \node[right=of gap2, xshift=\imageshift, yshift=\imageshift] (weather2) {\fbox{\includegraphics[trim={3px 3px 3px 3px},clip,width=\imagesize, height=\imagesize]{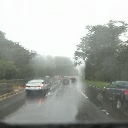}}};
    \node[right=of gap2, xshift=\imageshiftt, yshift=\imageshiftt] (weather3) {\fbox{\includegraphics[trim={3px 3px 3px 3px},clip,width=\imagesize, height=\imagesize]{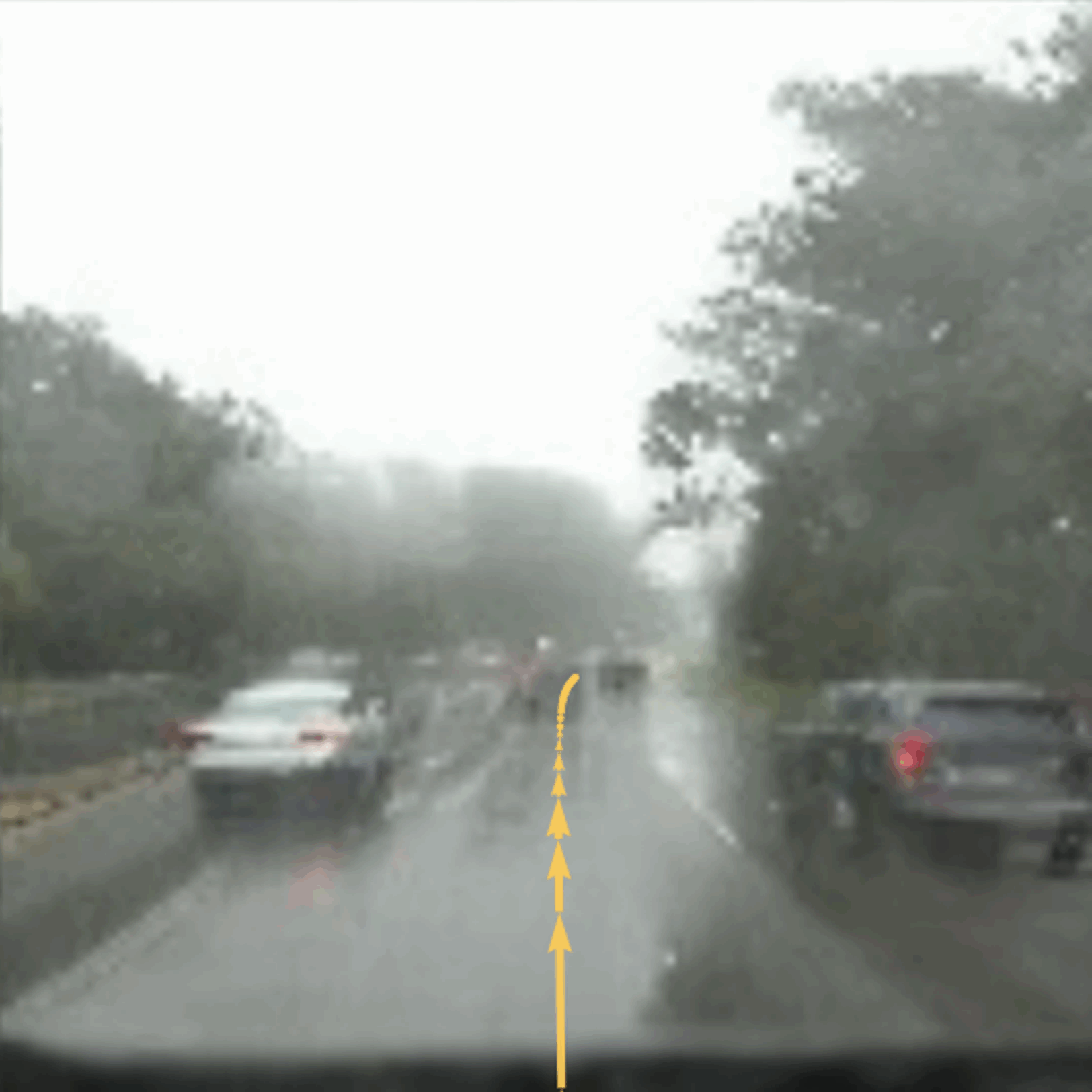}}};
    \node[right=of weather1, z=100] (time1) {\fbox{\includegraphics[trim={3px 3px 3px 3px},clip,trim={3px 3px 3px 3px},clip,width=\imagesize, height=\imagesize]{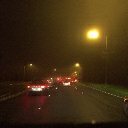}}};
    \node[right=of weather1, xshift=\imageshift, yshift=\imageshift] (time2) {\fbox{\includegraphics[trim={3px 3px 3px 3px},clip,width=\imagesize, height=\imagesize]{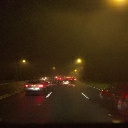}}};
    \node[right=of weather1, xshift=\imageshiftt, yshift=\imageshiftt] (time3) {\fbox{\includegraphics[trim={3px 3px 3px 3px},clip,width=\imagesize, height=\imagesize]{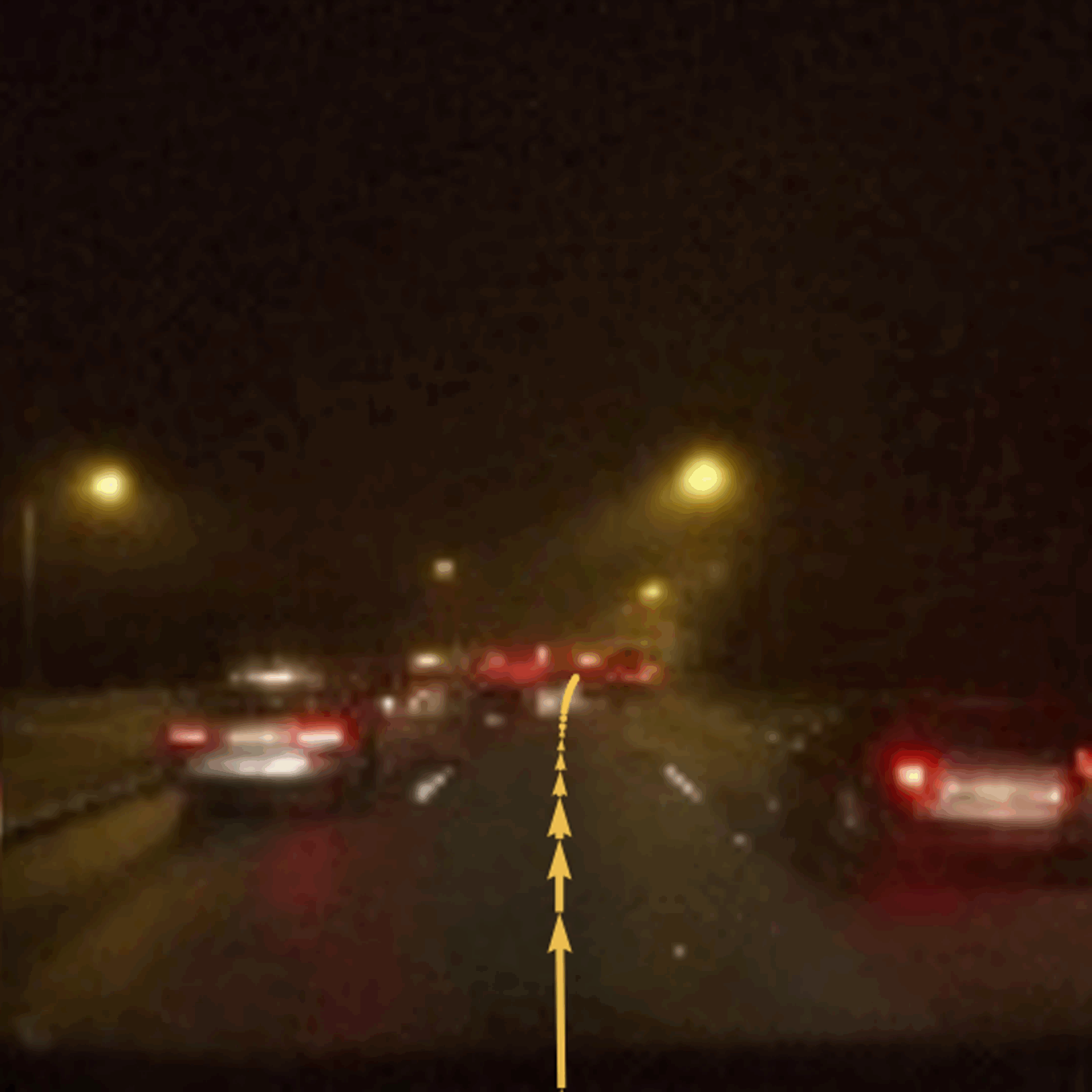}}};
    \node[right=of time1] (gap3) {};
    \node[right=of gap3, z=100] (augment1) {\fbox{\includegraphics[trim={3px 3px 3px 3px},clip,width=\imagesize, height=\imagesize]{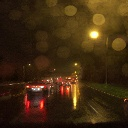}}};
    \node[right=of gap3, xshift=\imageshift, yshift=\imageshift] (augment2) {\fbox{\includegraphics[trim={3px 3px 3px 3px},clip,width=\imagesize, height=\imagesize]{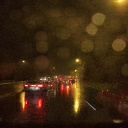}}};
    \node[right=of gap3, xshift=\imageshiftt, yshift=\imageshiftt] (augment3) {\fbox{\includegraphics[trim={3px 3px 3px 3px},clip,width=\imagesize, height=\imagesize]{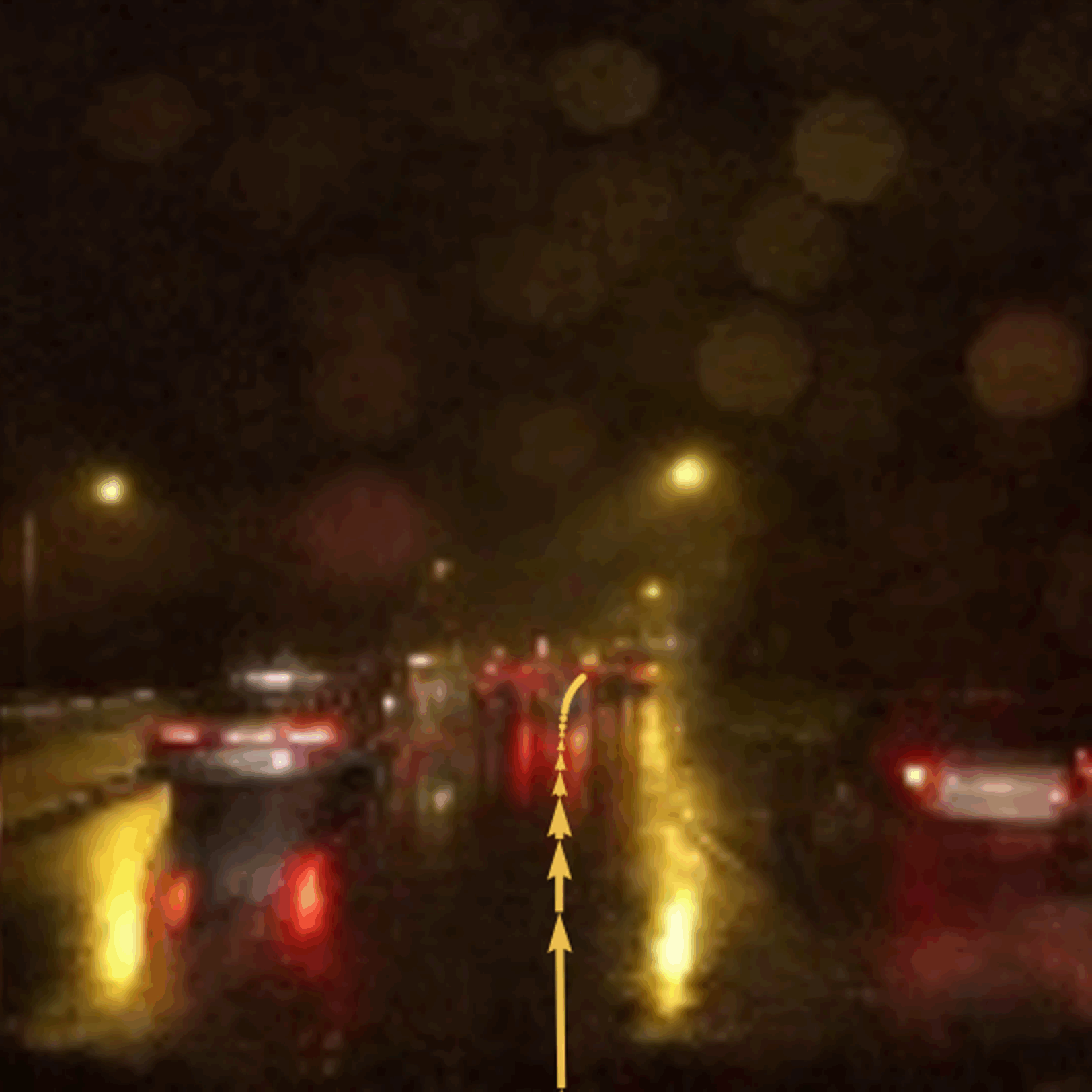}}};

    \node[below=\textgap of layout2] {Layout};
    \node[below=\textgap of log2] (logtext) {\textcolor{white}{g}Real\textcolor{white}{g}};
    \node[below=\textgap of same2, align=center] (sametext) {\textit{Same Conditions}};
    \node[below=\textgap of weather2, align=center] (weathertext) {\textit{Change Weather}};
    \node[below=\textgap of time2, align=center] (timetext) {\textit{Change Time}};
    \node[below=\textgap of augment2, align=center] (augmenttext) {\textit{Augment}};
    
    \node[draw, dotted, thick, rounded corners, waymogreen, fit=(log1) (log2) (log3) (same1) (same2) (same3) (logtext) (sametext)] {}; 
    \node[draw, dotted, thick, rounded corners, waymoblue, fit= (weather1) (weather2) (weather3) (time1) (time2) (time3) (weathertext) (timetext)] {}; 
    \node[draw, dotted, thick, rounded corners, waymopurple, fit=(augment1) (augment2) (augment3) (augmenttext)] {}; 

\end{tikzpicture}

    \vspace{-2em}
    \captionof{figure}{By connecting a driving video generation model with an end-to-end (E2E) planner, we can (1) \textcolor{waymogreen}{Evaluate Synthetic Data Quality via Planner} by controlling for the same traffic layout and scene conditions as the real videos to assess planner response discrepancies, (2) \textcolor{waymoblue}{Assess End-to-end Planner Domain Gap} via controlled experiments on operational conditions, and (3) \textcolor{waymopurple}{Improve E2E Planner Performance} on out-of-distribution domains via synthetic data from the video model. \textcolor{waymoamber}{Planner Predictions  ($\bm{\rightarrow}$)} overlaid. \textit{Generated data in italics.}}
    \label{fig:teaser}
\end{center}%
\end{strip}

\begin{abstract}

Recent advances in generative models have sparked exciting new possibilities in the field of autonomous vehicles.  Specifically, video generation models are now being explored as controllable virtual testing environments.  Simultaneously, end-to-end (E2E) driving models have emerged as a streamlined alternative to conventional modular autonomous driving systems, gaining popularity for their simplicity and scalability. However, the application of these techniques to simulation and planning raises important questions.  First, while video generation models can generate increasingly realistic videos, can these videos faithfully adhere to the specified conditions and be realistic enough for E2E autonomous planner evaluation? Second, given that data is crucial for understanding and controlling E2E planners, how can we gain deeper insights into their biases and improve their ability to generalize to out-of-distribution scenarios?
In this work, we bridge the gap between the driving models and generative world models (Drive\&Gen) to address these questions. We propose novel statistical measures leveraging E2E drivers to evaluate the realism of generated videos. By exploiting the controllability of the video generation model, we conduct targeted experiments to investigate distribution gaps affecting E2E planner performance. Finally, we show that synthetic data produced by the video generation model offers a cost-effective alternative to real-world data collection. This synthetic data effectively improves E2E model generalization beyond existing Operational Design Domains, facilitating the expansion of autonomous vehicle services into new operational contexts.

\end{abstract}    
\section{Introduction}
\label{sec:intro}
Autonomous vehicles (AV) promise to revolutionize transportation, but ensuring their safety and reliability remains a critical challenge. Typical AV development relies heavily on expensive and time-consuming real-world testing.  Recently, two promising technologies have emerged with the potential to transform AV development: end-to-end (E2E) driving models~\cite{hwang2024emmaendtoendmultimodalmodel, tian2024drivevlm} and video generation models~\cite{ho2022video, videoworldsimulators2024, gupta2023photorealisticvideogenerationdiffusion}. E2E models offer a simplified approach to AV control by directly mapping sensor input to planning output, enabling the simplification of the AV stack and model scaling. On the other hand, video generation models can generate realistic sensor data for testing and training.

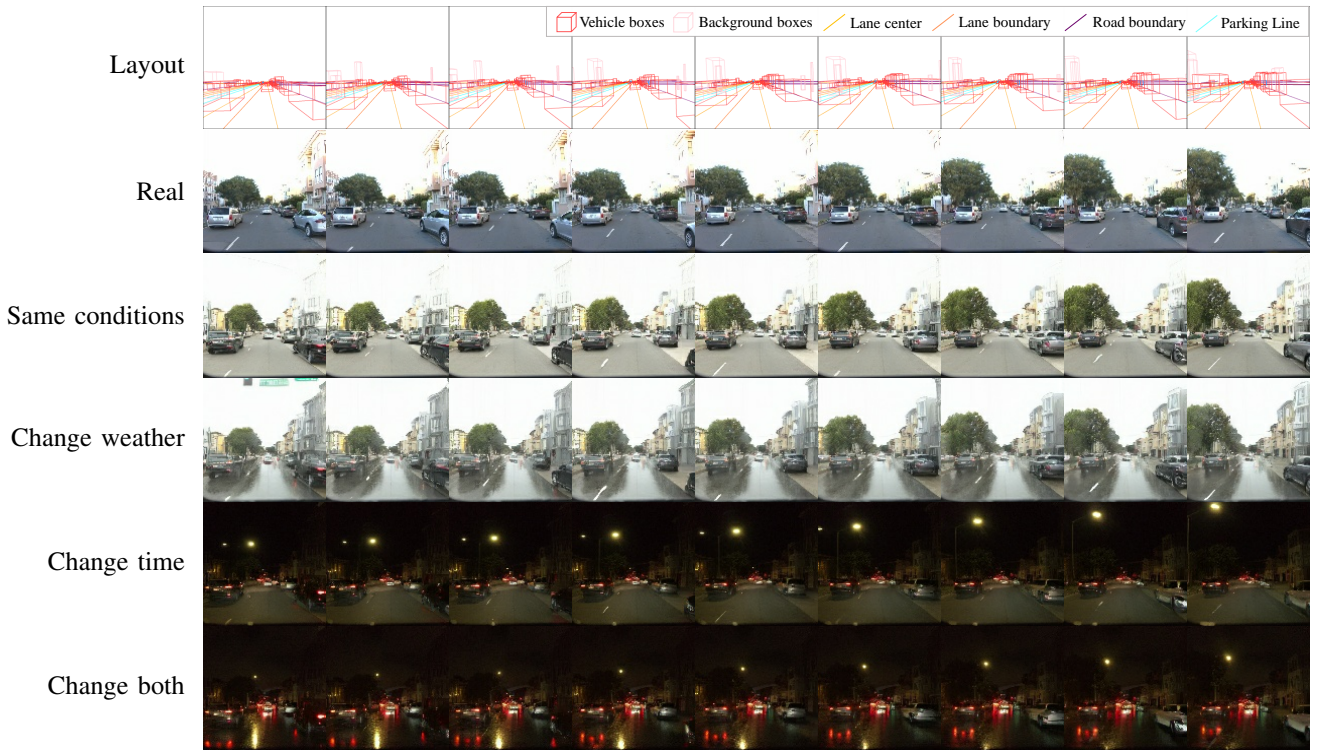
\begin{figure*}
\centering

\begin{tikzpicture}

\def\imagewidth{14.7cm}
\def\imageheight{1.8cm}
\def\verticalgap{-0.23cm} %

\node (image1) {\includegraphics[width=\imagewidth]{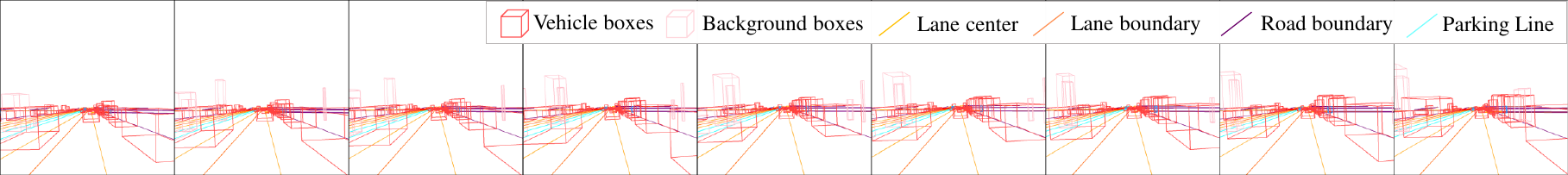}}; 
\node[anchor=east] at (image1.west) {Layout}; 

\node[below=\verticalgap of image1] (image2) {\includegraphics[width=\imagewidth]{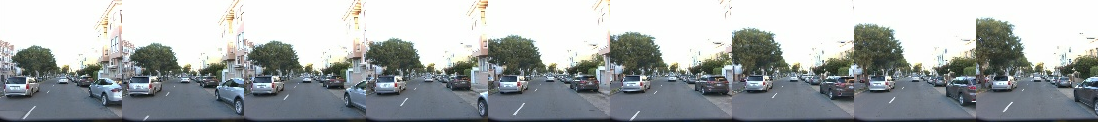}}; 
\node[anchor=east, align=right] at (image2.west) {Real}; 

\node[below=\verticalgap of image2] (image3) {\includegraphics[width=\imagewidth]{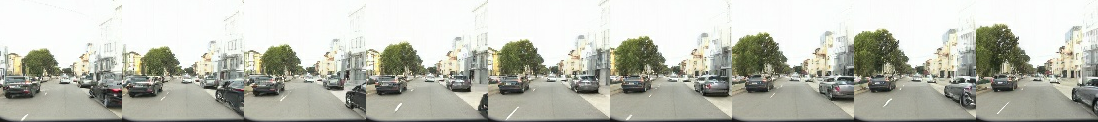}}; 
\node[anchor=east, align=right] at (image3.west) {Same conditions}; 

\node[below=\verticalgap of image3] (image4) {\includegraphics[width=\imagewidth]{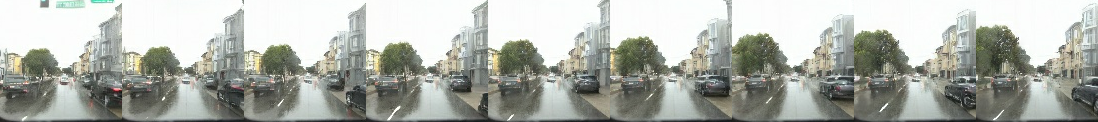}}; 
\node[anchor=east, align=right] at (image4.west) {Change weather};

\node[below=\verticalgap of image4] (image5) {\includegraphics[width=\imagewidth]{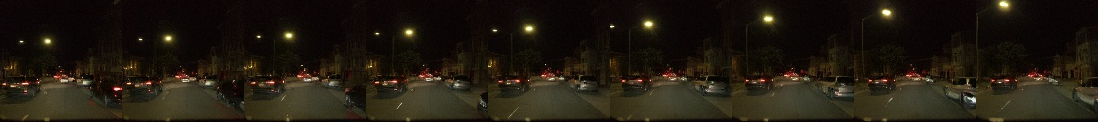}}; 
\node[anchor=east, align=right] at (image5.west) {Change time};

\node[below=\verticalgap of image5] (image6) {\includegraphics[width=\imagewidth]{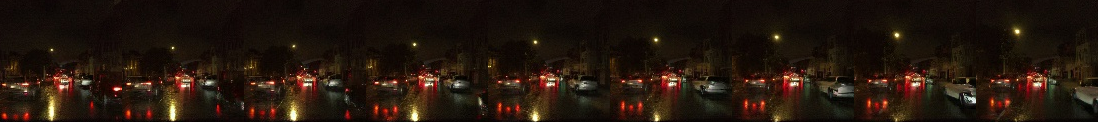}}; 
\node[anchor=east, align=right] at (image6.west) {Change both};

\end{tikzpicture}
\caption{Generated videos conditioned on various conditions. (1) The top row displays the input conditions, including road maps and bounding boxes, projected to the camera. (2) The second row shows the corresponding real-world video. The subsequent rows demonstrate the model's ability to generate videos under different conditions: (3) identical conditions to the original video, (4) changing the weather from no-rain to rain, (5) changing the time of day to 00:00 (at midnight), (6) with both rain and nighttime conditions.}
\label{fig:videogen}

\end{figure*}

Despite their potential, key questions remain for these technologies. While recent work on generating synthetic driving videos have shown increasingly impressive visual quality, it remains unclear if that correlates with the planner's response. As shown in the adversarial literature~\cite{goodfellow2014explaining}, even the slightest perturbation in the image that is barely visible to the human eye can result in a dramatically different output response of a downstream deep learning model (e.g., predicting a panda to be a baboon). How planning models perceive the realism gap between real and synthetic data remains an open question. To our knowledge, we are among the first works to study the realism of such video generation model to facilitate the development and evaluation of an end-to-end planner.

Meanwhile, E2E planner models present a different set of challenges. While E2E models greatly simplify the model formulation by directly mapping sensor inputs to controls, 
it poses a key challenge on how to evaluate such models, especially their performance on out-of-distribution domains.

To address these questions, our \textit{key observation} is that for a certain driving scene, the expected driving behavior should largely be a result of the underlying traffic scene layout (e.g., road map layout and agent features such as locations, types, and sizes) and mostly independent of other visual features, such as lighting conditions, weather conditions and the appearance of each agent (e.g., red vs blue car). This is the core underlying assumption in all behavioral simulation tasks \cite{Montali23neurips_wosac}. A video generation model, conditioned on both the scene layout and visual features such as weather and time-of-day, can generate the same underlying traffic scenario under different visual conditions.

In this light, we present \ours{}, a framework for co-evaluating E2E driving models and video generation models (see Fig.~\ref{fig:teaser}). First, by controlling for the same scene layout and visual conditions as the real videos, we can study the responses from the same end-to-end planner model based on each real scene and its synthetic counterpart to evaluate the sim-to-real domain gap of the video generation model. We introduce novel statistical measures utilizing E2E driver behavior within the generated environments to quantify the realism of these virtual worlds. Second, due to the ability of the controllable video generation models to generate traffic scenarios of the \emph{same} layout and \emph{different} operational design domains (ODD) such as varied weather and time-of-day, we are able to do \emph{controlled experiments} to evaluate E2E planner performance under varied ODDs for model diagnostics and new ODD expansion readiness assessment. Finally, we demonstrate that synthetically generated data can be an effective mechanism to improve out-of-distribution generalization of E2E planner models.

In summary, the main contributions of this work are:
\begin{itemize}
\item Introduces novel statistical measures for evaluating the realism of video generation models from the perspective of E2E driving models.
\item Analyzes the performance differences of the E2E planner in in-distribution versus out-of-distribution contexts.
\item Demonstrates the effectiveness of synthetic data generated by the video generation model for improving E2E model generalization to out of distribution scenarios.
\end{itemize}

\section{Related Work}
\label{sec:related-work}

\noindent \textbf{World Models.}
World models~\cite{LeCun2022world} refer to learned representations of the environment and its dynamics. During early explorations, it has showcased remarkable success in various applications~\cite{NEURIPS2018_2de5d166, hafner2019dreamer, hafner2021mastering, hafner2023dreamerv3, Kim2020_GameGan}. Constructing world models in real-world driving settings poses unique challenges because of the high sample complexity in driving worlds.  Recently, with the development of diffusion-based video generation~\cite{ho2022video, singer2023makeavideo, wu2023tune, blattmann2023videoldm, blattmann2023stablevideodiffusionscaling, Peebles2022DiT, videoworldsimulators2024, gupta2023photorealisticvideogenerationdiffusion}, world models~\cite{hu2023gaia1generativeworldmodel, ma2024unleashinggeneralizationendtoendautonomous, wang2023drivedreamerrealworlddrivenworldmodels, wen2024panacea, wang2024driving, gao2024vista, li2023drivingdiffusion, yang2024genad, gao2023magicdrive} are capable of generating photo-realistic videos, conditioning on user controls. 
GAIA-1~\cite{hu2023gaia1generativeworldmodel} and Vista~\cite{gao2024vista} generate the future world with video diffusion models~\cite{ho2022video, blattmann2023stablevideodiffusionscaling} conditioning on text prompts and driving actions.  DriveDreamer~\cite{wang2023drivedreamerrealworlddrivenworldmodels}, DrivingDiffusion~\cite{li2023drivingdiffusion}, MagicDrive~\cite{gao2023magicdrive}, and Panacea~\cite{wen2024panacea} further generate controllable multi-view videos~\cite{nuscenes}. 
Concurrent work Delphi~\cite{ma2024unleashinggeneralizationendtoendautonomous} also uses world models to improve E2E driving models, but our method more comprehensively covers all the stages of AV software development including evaluation, ODD-specific performance analysis, and synthetic data augmentation~\cite{Mumuni_2024, bowles2018gan, ma2023generating}.

\noindent\textbf{End-to-end Planning Models.}
For E2E driving, \cite{xu2017end} introduces a method to capture the temporal sequence of visual inputs, enabling direct learning from driving videos. 
Recently, end-to-end (E2E) driving has garnered increasing attention. Some approaches~\cite{hu2022st, hu2023planning,casas2021mp3,gu2023vip3d,jiang2023vad,sadat2020perceive} enable gradient back-propagation across modules, enhancing inter-task communication and mitigating error accumulation. Another line of research leverages pre-trained vision-language models (VLMs), which embed common-sense knowledge acquired from large-scale Internet data~\cite{tian2024drivevlm,wang2024omnidrive,hwang2024emmaendtoendmultimodalmodel,bai20243dtokenizedllmkeyreliable}. Pioneer work DriveVLM~\cite{tian2024drivevlm} uses VLMs and chain-of-thought~\cite{wei2022chainofthoughtpromptingelicitsreasoning} prompting to describe critical objects and produce hierarchical planning signals, including high-level decisions and low-level waypoints. While these work mostly focus on image-only setting,  Atlas~\cite{bai20243dtokenizedllmkeyreliable} integrate 3D signals into VLM and showed strong results in both perception and planning. 
In this work, we also leverage the pre-trained vision-language model PaLI~\cite{chen2023pali}, to build our end-to-end planning model.

\noindent\textbf{Planner and World Model Evaluation.}
Evaluation of E2E planners in existing literature typically involves open-loop and closed-loop metrics. Open-loop evaluation measures how closely the planner’s predictions match ground-truth labels when the planner is not interacting with a dynamic environment~\cite{chen2024end}. In a closed-loop context, the E2E planner operates within either a simulator or a real-world environment. While certain benchmarks~\cite{jia2024bench2drive} provide closed-loop simulation capabilities, concerns persist regarding the validity of these simulators and the realism of their synthetically generated sensor data. Existing studies~\cite{ma2024unleashinggeneralizationendtoendautonomous} generally do not directly evaluate the realism of synthetic data; instead, they use it as additional training material during fine-tuning and report the resulting performance gains. In this paper, we propose a novel framework that directly measures simulation realism.

\section{Method}
The primary focus of this work is introducing a novel co-evaluation framework for video generation and E2E driving, which is introduced in Sec.~\ref{sec:coeval}.
In Sec.~\ref{sec:videogen}, we describe how the diffusion-based video generative model is built especially how we encode various control modalities, including bounding boxes, road maps, ego-car pose, time-of-day, and weather. 
In Sec.~\ref{sec:planner}, we provide details on how we extend a pre-trained vision-language model (VLM) into an E2E planner.

\subsection{Co-evaluation}
\label{sec:coeval}
Traditional metrics for evaluating video generation cannot fully capture visual quality and controllability~\cite{stylegan_v}. Moreover, isolating factors like traffic, weather, and time-of-day is costly in real-world data and demands precise control in synthetic environments. These issues motivate our proposed co-evaluation framework, which systematically measures both video generation quality and planning performance in diverse scenarios. To evaluate the video generation, we feed the generated videos into an E2E planner and compare how closely the planner’s responses match those observed in real scenes with an equivalent layout. By adjusting the conditions for the video generation model (e.g., weather or time-of-day), we can further analyze the planner’s behavior under different scenarios and track performance changes.

We first introduce widely used metrics for E2E planning (Average Displacement Error) and video generation (Fr\'echet Video Distance), then present our Behavior Permutation Test.

\noindent\textbf{Average Displacement Error (ADE).} ADE measures the mean L2 distance between predicted and ground-truth trajectories, typically calculated at future horizons of 1s, 3s, and 5s. Although ADE offers a straightforward comparison, it is not highly discriminative. For instance, two trajectories with the same ADE can deviate in opposite directions, yet exhibit fundamentally different errors.

\noindent\textbf{Fr\'echet Video Distance (FVD).}  FVD~\cite{unterthiner2018towards} is a widely used metric that correlates well with human perception of photo-realism. It measures the distributional distance between real and generated videos in a latent feature space. In our evaluation, we randomly sample 5,000 videos from both the logged dataset and synthetic outputs to compute FVD. However, FVD alone does not fully capture whether the generated video adheres to the conditions.

\noindent\textbf{Behavior Permutation Test (BPT).} We propose \emph{BPT}, a novel metric to assess whether generated videos can ``fool'' the planner, by measuring how similarly the planner responds ``behaviorally'' to the generated scene versus the real scene.

For each driving scene, we conduct a permutation test as follows. 
We feed the planner with real data and sample $M$ planned trajectories, denoted as 
$\{ \tau_{i}^{\text{real}} \}_{i=1}^M$, where $\tau_i \in \mathbf{R}^{q\times 2}$ is the way-point representation of trajectory and $q$ is the number of points.
Similarly, we feed the planner with data generated by the video generation model under the same conditions, 
and sample $N$ planned trajectories 
$\{ \tau_{j}^{\text{gen}} \}_{j=1}^N$. In the experiments, we use $M = N = 10$.

The null hypothesis posits that both sets of trajectories originate from the same distribution. Formally,
\begin{equation}
    H_0 : \{ \tau_{i}^{\text{real}} \}_{i=1}^M \overset{d}{=} \{ \tau_{j}^{\text{gen}} \}_{j=1}^N
\end{equation}

The test statistic \( T \) for the permutation test is a generalized version of Chamfer distance between the two sets of trajectories, denoted \( D(\cdot) \). Formally, 
\begin{align}
    T &= D\left(\{\tau_i\}_{i=1}^{M}, \{\tau_j\}_{j=1}^{N}\right)\label{eq:def} \\
    &= \frac{1}{2M} \sum_{i=1}^M \min_{j \in [1, N]} \left\| \tau_i - \tau_j \right\|_2  + \frac{1}{2N} \sum_{j=1}^N \min_{i \in [1, M]} \left\| \tau_i - \tau_j \right\|_2 \nonumber 
\end{align}

For \( n = 1000 \) times, we randomly permute the trajectories and create two new trajectory sets $\{\tau_{i'}\}$ of size $M$ and $\{\tau_{j'}\}$ of size $N$, and recalculate the test statistic \( T' \) using Eq.~\ref{eq:def}. Denote $T_0$ to be the distance when we have one set that consists of only real trajectory (i.e. $\{ \tau_{i}^{\text{real}} \}_{i=1}^M$) and one set consists of only generated trajectory (i.e. $\{ \tau_{j}^{\text{gen}} \}_{j=1}^N$). We compute probability \( \text{P}(T' > T_0)\) to be the \( p \)-value for each scene, representing the probability that the observed difference between trajectory sets is solely due to random chance. Specifically, \( p < 0.05 \) indicates that the video generation model \emph{fails} the Behavior Permutation Test, implying that the planner behaves significantly differently when fed real versus generated data.

\subsection{Video Generation Model} 
\label{sec:videogen}

\begin{figure}[t]
  \centering
   \includegraphics[width=3.28in]{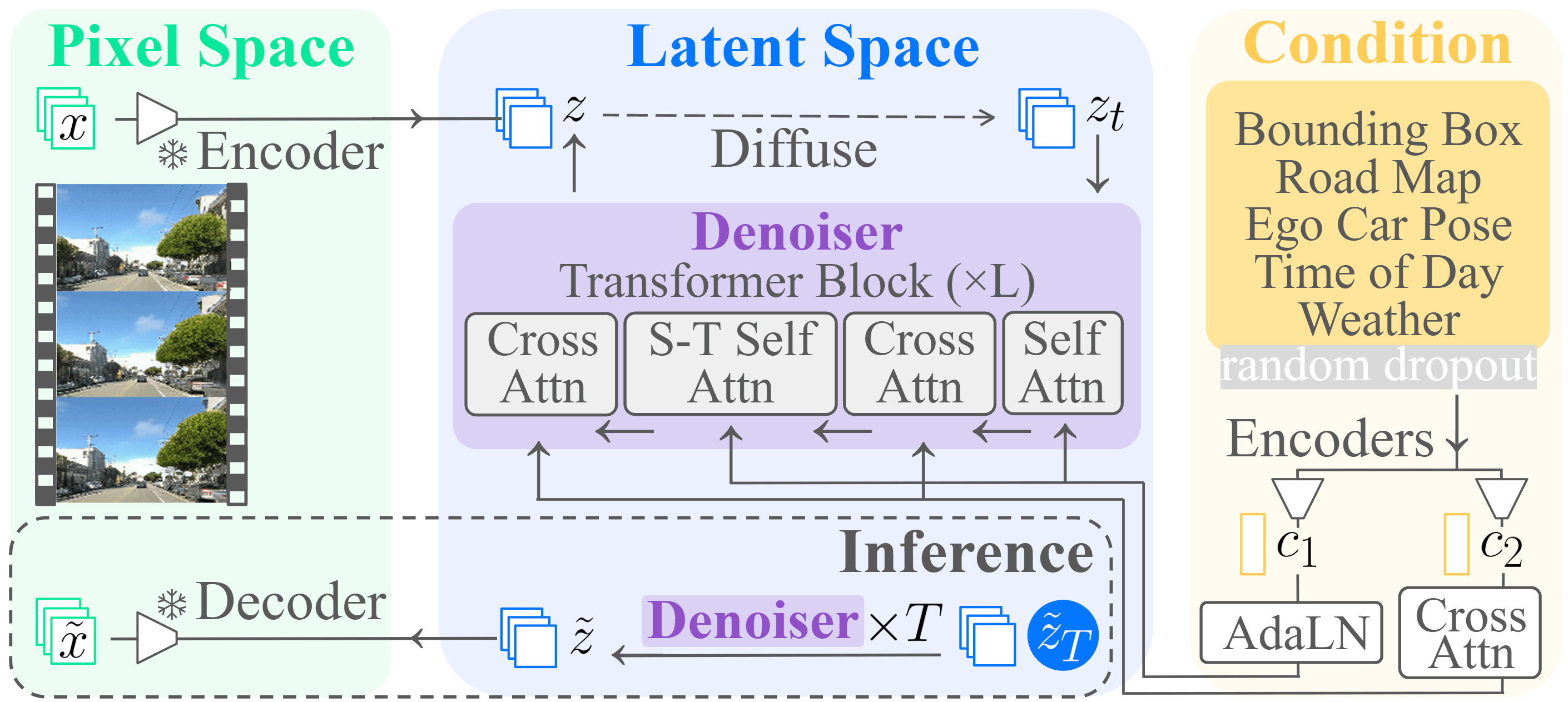}
   \caption{Model architecture of our video generation model. We enable control of scene and traffic layout (bounding boxes, road map, and ego car pose) and operational conditions (time-of-day, weather), extending the latent video diffusion model W.A.L.T \cite{gupta2023photorealisticvideogenerationdiffusion}. The conditions are encoded and interact with intermediate features in the diffusion transformer via a combination of AdaLN and cross attention mechanisms. The model is fine-tuned on a large corpus of driving videos.}
   \label{fig:model}
\end{figure}

We develop a controllable video diffusion model based on the pre-trained W.A.L.T ~\cite{gupta2023photorealisticvideogenerationdiffusion}. We extend it by enabling additional control modalities derived from real-world driving data, including bounding boxes, road map, ego car pose, time-of-day, and weather. This enables us to generate videos that are not only visually realistic but also adhere to specific driving scenarios, providing a more controllable video generation framework. We introduces two novel designs: (1) Fine-grained time control: recise manipulation of time-of-day, enabling smooth transitions between various lighting conditions. (2) Efficient condition representation: a sparse 3D representation for bounding boxes and road maps via learned tokenization, significantly reducing memory consumption. 
The overall architecture is shown in Fig.~\ref{fig:model}.

\noindent\textbf{Bounding box.} Each bounding box is represented as a 8-dimensional vector consisting of position $(x, y, z)$, dimensions (width, height, length), yaw angle, and type. We encode the yaw angle using sinusoidal functions, and process the box types using one-hot embeddings. An MLP projects this vector into a 256-dimensional space. To handle the varying numbers of bounding boxes, we set a max number of 256 for each frame and apply padding or truncate as needed. We transform the bounding boxes from world coordinate into an ego-vehicle coordinate system.

\noindent\textbf{Road maps.} Following~\cite{nayakanti2023wayformer}, road maps are represented as line segments. We set the max number of line segments as 4,096. Each line segment has 3 attributes, i.e., starting point position, ending point position, and type. We transform the positions into an ego-vehicle coordinate system as for the bounding boxes. Segment types are encoded into one-hot vectors. We project the segment features into a 256-dimensional space using an MLP, and reduce the number of tokens by using a latent query attention~\cite{jaegle2021perceiver, nayakanti2023wayformer}, which reduces computation and memory utilization.

\noindent\textbf{Ego-car pose.} The pose of the ego-vehicle is flattened into a 12-dimensional vector, comprising a 3$\times$3 rotation matrix and a 3-dimensional translation. This vector is then projected into a 256-dimensional space using an MLP.

\noindent\textbf{Time-of-Day.} We enable precise control of time-of-day, allowing for specific time inputs such as ``06:41" or ``20:25". Since the same time-of-day can have very different lighting condition in different seasons or in different geographic locations, we propose to use sun angles instead. We use solar azimuth $\theta$ and elevation $\phi$ angles, which can be calculated from the local time-of-day $t_d$, time of year $t_y$, and geographic location $l_{geo}$ (latitude, longitude). By manipulating the time-of-day $t_d$ given certain $t_y$ and $l_{geo}$, we get different sun angles $(\theta, \phi)$ and generate videos with diverse lighting scenarios based on $(\theta, \phi)$. The sun angles are then encoded using sinusoidal functions of different frequency, and projected into a latent space via MLP.

\noindent\textbf{Weather.} Weather conditions, such as rain or no-rain, are encoded using a one-hot vector and then also projected to a latent space using an MLP.

We concatenate the embeddings of all these conditions into a unified sequence $z$, and pass through a transformer encoder to get feature $f_{z}$. $f_{z}$ is then incorporated into the diffusion model through cross-attention, enabling effective conditioning. Additionally, similar to~\cite{gupta2023photorealisticvideogenerationdiffusion} we employ a pooled representation of the feature $f_{z}$, processed through an MLP, to modulate the multi-head attention and feed-forward layers in the diffusion model's self-attention blocks. This mechanism allows for adaptive scaling and shifting of feature representations, leading to more precise control over the generated video content.

\subsection{End-to-end Driving Model}
\label{sec:planner}
We train the E2E driving model based on a pretrained vision-language model PaLI~\cite{chen2023pali} following EMMA~\cite{hwang2024emmaendtoendmultimodalmodel}. To efficiently handle temporal frames, we adopt a collaged-image representation~\cite{shi2023zero123++}, arranging a 3$\times$3 grid of images from left to right, top to bottom, with the earliest frame in the top-left corner. This collage is encoded into 1,536 tokens and concatenated with text tokens before being processed by the encoder-decoder model. The input text includes additional data such as the self-driving car’s past states (e.g., position and velocity) and routing instructions (e.g., ``turn left", ``go straight").
The model is trained to generate trajectories from temporal frames, where each trajectory is represented as a sequence of waypoints encoded as float values in text format, framing the planning task as a Visual Question Answering (VQA) problem.
We initialize our model with pre-trained weights and fine-tune the language decoder, keeping the vision encoder fixed. The model is trained using cross-entropy loss.

\section{Experiments}
\label{sec:exp}
In this section, we first describe the model training and dataset details. Sec.~\ref{sec:exp_videogen} evaluates the controllable video generation model and shows the model's ability to generate videos that closely align with the specified conditions.
We also assess the similarity between the real and generated videos using BPT.
Subsequently, in Sec.~\ref{sec:exp_planner}, we leverage the versatility of the video generation model to create diverse driving scenarios and test the E2E planner.
In Sec.~\ref{sec:exp_improve}, we demonstrate that our high-quality synthetically generated data improve the performance of the E2E planner.
Since this work focuses on a novel co-evaluation framework rather than state-of-the-art video generation, detailed FVD or resolution comparisons lie beyond our scope. Moreover, current methods lack the fine-grained control (e.g., minute-level time-of-day/sun angles) required for comprehensive planner evaluation. Finally, UniAD’s~\cite{hu2023planning} deterministic trajectory prediction is incompatible with the proposed BPT, thus cannot be used in our framework.

\noindent\textbf{Model Training.}
To achieve video generation with conditions, we curated a dataset with 
about ten million driving segments, among which we hold out 1\% for testing and use the remaining for training.
Each segment includes 17 frames in 10 Hz with a resolution of $128\times128$ pixels, and comes with multiple features including agent bounding boxes, road map, ego-car trajectory, local time, geo-location and weather. We use a maximum of 256 bounding boxes per frame. We train our video generation model on this for 700k steps with a batch size of 64. During training, we randomly dropout each condition with the probability of 0.1. This improves generalization for the models and allows us to run inference without some of the conditions. We fine-tune the VLM-based E2E planner for 120k steps.

\subsection{Evaluation of Controllable Video Generation}
\label{sec:exp_videogen}

\newcommand{\cwidth}{4.3em}  %
\newcolumntype{C}[1]{>{\centering\arraybackslash}p{#1}} 

\begin{figure}
\centering
\setlength{\tabcolsep}{0pt}
\begin{tabular}{@{}cC{\cwidth}C{\cwidth}C{\cwidth}C{\cwidth}C{\cwidth}@{}}
\toprule
\centering & Real & \scalebox{0.9}{Same Cond.} & w/o Box & Rain & Night \\
\midrule
 &

\includegraphics[trim={3px 3px 3px 3px},clip,width=0.98\linewidth]{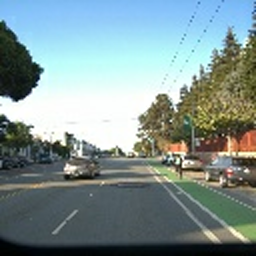} & \includegraphics[trim={3px 3px 3px 3px},clip,width=0.98\linewidth]{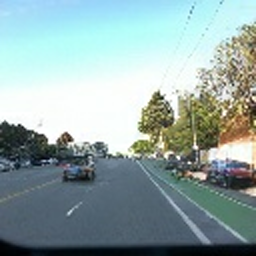} & \includegraphics[trim={3px 3px 3px 3px},clip,width=0.98\linewidth]{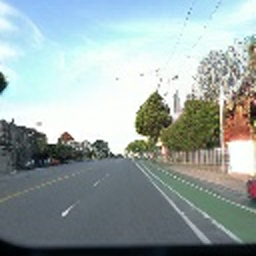} & \includegraphics[trim={3px 3px 3px 3px},clip,width=0.98\linewidth]{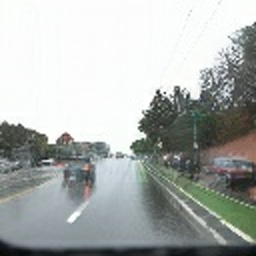} & \includegraphics[trim={3px 3px 3px 3px},clip,width=0.98\linewidth]{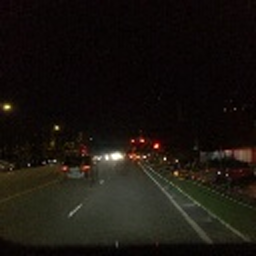} \\
\midrule
FVD & - & 39.89 & 38.97 & 151.25 & 493.37 \\
ADE & 0.7548 & 0.8594 & 1.1216 & 0.8736 & 0.8760 \\
BPT & - & 69.62\% & 55.28\% & 69.28\% & 67.66\% \\
\bottomrule
\end{tabular}

\caption{
\footnotesize Evaluation of controllable video generation with FVD, ADE@5s, and BPT on 5000 random samples. FVD doesn't fully capture visual quality -- FVD for Rain/Night (relatively rare in our dataset) are much higher (because of distribution shifts) though the photo-realism of videos are visually similar. FVD cannot measure controllability -- removing the conditioning on bounding boxes greatly changes the car locations but has little effect on FVD. ADE and BPT don't suffer from such data distribution shifts, and can capture model controllability -- both metrics are notably worse when bounding boxes are removed.}
\label{fig:fvd_ade_bpt}
\end{figure}

We evaluate the realism of the generated videos and consider a few candidate metrics. A commonly considered video realism metrics is the FVD score \cite{unterthiner2018towards}. However, as this is a distributional matching metric, it measures distributional differences and not necessarily visual quality. In Fig. \ref{fig:fvd_ade_bpt}, we show that our FVD for night-time driving is disproportionately worse than the videos generated with the same driving conditions as the logged data though their visual realism is on par.

An alternative for directly measuring video quality is to measure the resulting planner performance by ADE. However, though ADE is a good measurement of planner quality, it doesn't indicate whether generated videos, conditioned on the same traffic layout, elicits a similar planner prediction. That is because two vastly trajectory outputs (one leaning left, one leaning right) could end up with similar ADE compared to ground truth. A higher ADE could also be due to worse planner performance in certain operational conditions (rain, night), and not necessarily unrealistic video inputs. In other words, the ADE metric doesn't allow us to easily disentangle the performance of the video generation model, versus the E2E planner itself.

\begin{figure}[t]
  \centering
    \includegraphics[width=0.478\textwidth]{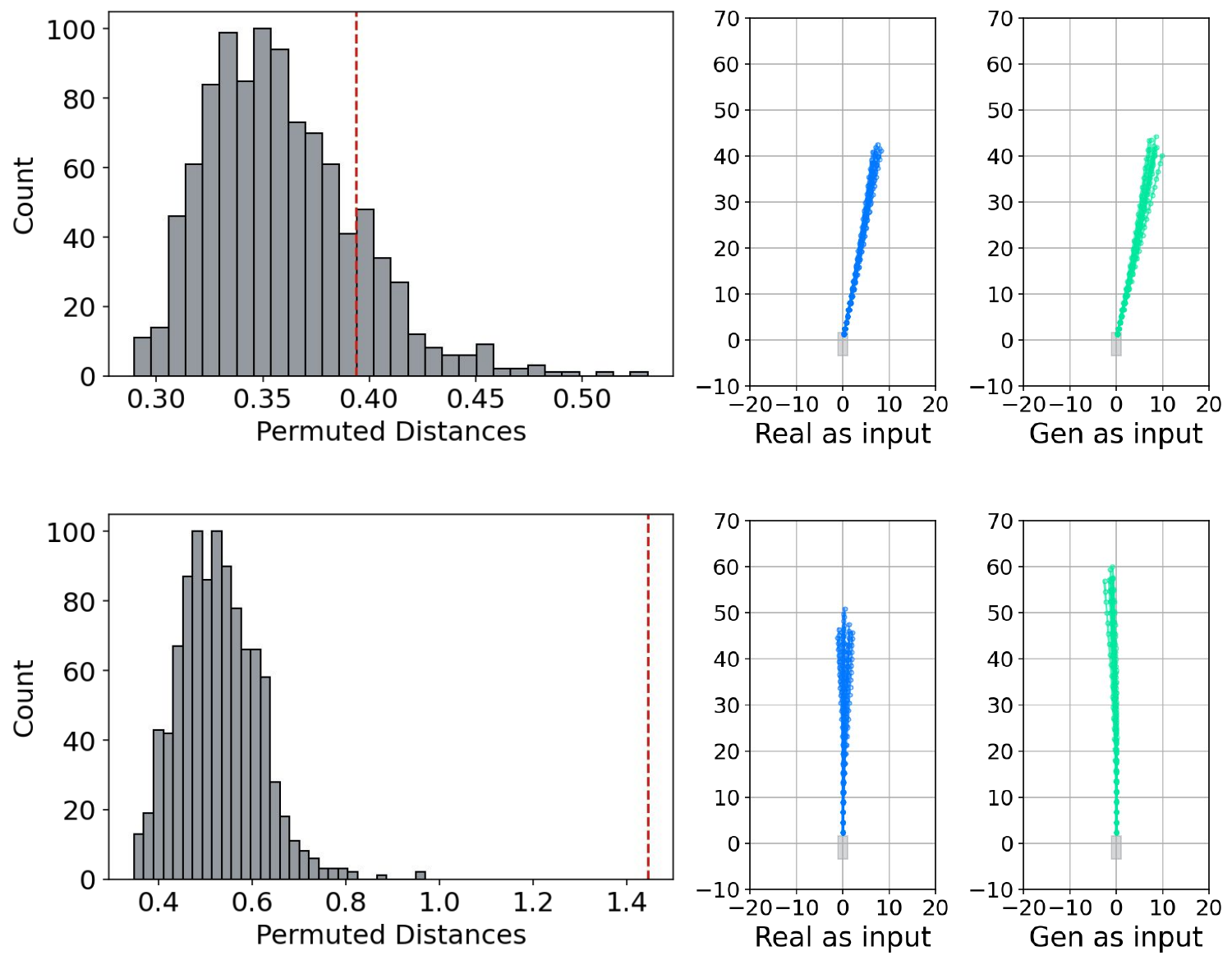}
   \caption{Behavioral Permutation Test (BPT) visualizations. BPT performs a set-to-set comparison of predicted trajectories from real and generated videos.
    In the top row, when the two sets of trajectories are similar, the distance between the two sets (red dash line) falls well within permuted distributions, resulting in a failure to reject the null hypothesis. The bottom shows a rejection of the null hypothesis,  where the two sets of trajectories are significantly different from each other.
    }
   \label{fig:hypothesis}
\end{figure}

Finally, we consider the Behavior Permutation Test (BPT) metric, as introduced in Sec. \ref{sec:coeval}. In Fig. \ref{fig:hypothesis} we demonstrate a pair of failure-to-reject and rejection examples. When the two sets of trajectories are similar, the distance between the original two sets (red dash line) falls well within permuted distributions while it significantly falls out-of-distribution when the two sets of trajectories are significantly different. For each scene, BPT emits signals for whether trajectory plans from real v.s. synthetic data are sufficiently similar. We measure the fail-to-reject rate of BPT over the entire validation set to obtain an average. Importantly, note that the expected ceiling for the BPT fail-to-reject rate is 95\% (the nominal confidence level), since the hypothesis test rejects all cases with $p < 0.05$.

In this light, we evaluate the quality of our video generation model. Qualitative results can be found in Fig. \ref{fig:videogen} and more quantitative results in Fig. \ref{fig:fvd_ade_bpt}. Conditioned on the same conditions as real data, we obtain $69.62\%$ BPT failure-to-reject rate (out of $95\%$ expected ceiling), indicating a broadly similar planner response when presented with real and synthetic data. We sanity check that altering the scene layout (removing bounding box constraints) leads to a steep drop in BPT failure-to-reject rate, while modifying operational conditions (rain/night) result in small but not-insignificant planner behavior changes, due to varied performance of the planner under different operational conditions, which we further investigate in Sec. \ref{sec:exp_planner}. 

We show an ablation study assessing the effectiveness of using sun angles for time-of-day encoding in Table~\ref{tab:vid_compare}. By comparing our model to a baseline that uses local time, we demonstrate better performance in FVD, ADE, and BPT. This suggests that sun angle encoding provides a more informative representation of time-of-day variations, leading to more realistic and controllable video generation.

\begin{table}
  \caption{Comparison of video generation models. This table compares our model with a baseline conditioned on local time instead of sun angles. The results highlight how sun angle encoding yields more realistic and controllable videos, reflected by improvements in FVD, ADE, and BPT.}
  \label{tab:vid_compare}
  \centering
  \begin{tabular}{@{}p{3.2cm}ccc@{}}
    \toprule
    Time-of-day encoding & FVD & ADE@5s & BPT \\
    \midrule
    Local time & 45.54 & 0.8739 & 68.46\% \\
    Sun angles (\textbf{ours}) & \textbf{39.89} & \textbf{0.8594} & \textbf{69.62\%} \\
    \bottomrule
  \end{tabular}
\end{table}

\subsection{Evaluation of End-to-End Planner}
\label{sec:exp_planner}

By leveraging controllable video generation, we can systematically manipulate various conditions, such as weather and time-of-day, to create diverse and realistic driving scenarios. This enables us to isolate the impact of individual factors on planner behavior, leading to a deeper understanding of the model's strengths and weaknesses. For instance, by generating videos with varying levels of illumination, we can assess the planner's performance under different lighting conditions, without the confounding effects of driving mix shifts, such as reduced traffic density at night (usually associated with better planner performance). Our model allows for precise control over individual conditions, enabling a more granular analysis of the planner's behavior. To evaluate the planner's performance under these controlled conditions, we employ ADE, which directly measures the discrepancy between predicted and ground-truth trajectories.

\begin{table}
  \centering
  \caption{ADE scores on real and generated videos. Removing scene layout conditions (bounding box and road map)  significantly increases the ADE, while removing operational conditions (weather and time-of-day) has a less pronounced impact on ADE.}
  \label{tab:ade-log-gen}
  \begin{tabular}{@{}lccc@{}}
    \toprule
    Input videos & ADE@1s & ADE@3s & ADE@5s \\
    \midrule
    Real & 0.0288 & 0.2606 & 0.7548 \\
    Gen & 0.0300 & 0.2859 & 0.8594 \\
    \hdashline
    Gen w/o bbox & 0.0437 & 0.3814 & 1.1216\\
    Gen w/o road map & 0.0348 & 0.3059 & 0.9111\\
    \hdashline
    Gen w/o weather & 0.0299 & 0.2857 & 0.8593\\
    Gen w/o time-of-day & 0.0299 & 0.2886 & 0.8751\\
    \bottomrule
  \end{tabular}

\end{table}

\begin{figure}
    \centering
    \includegraphics[width=0.236\textwidth]{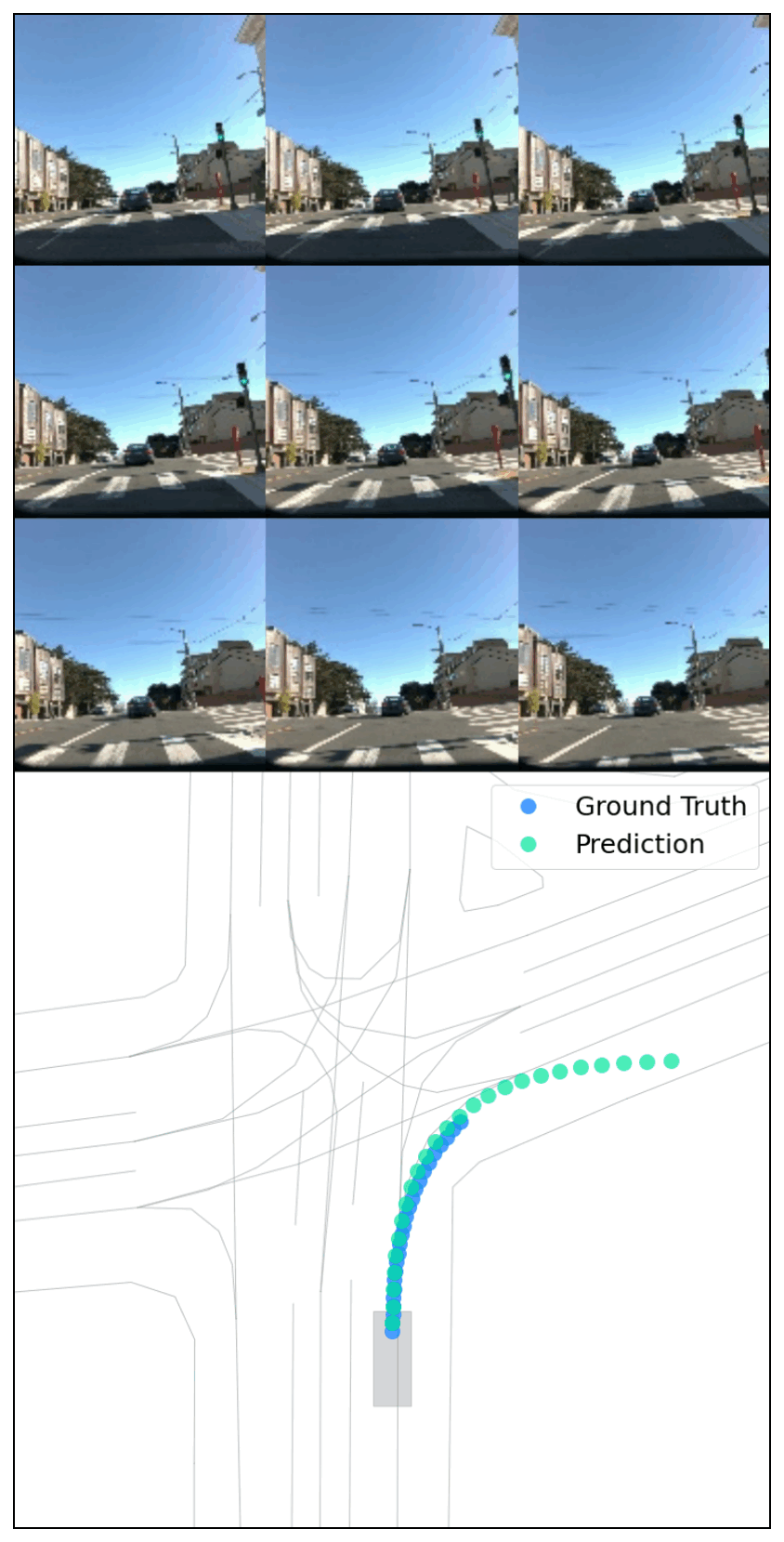}
    \includegraphics[width=0.236\textwidth]{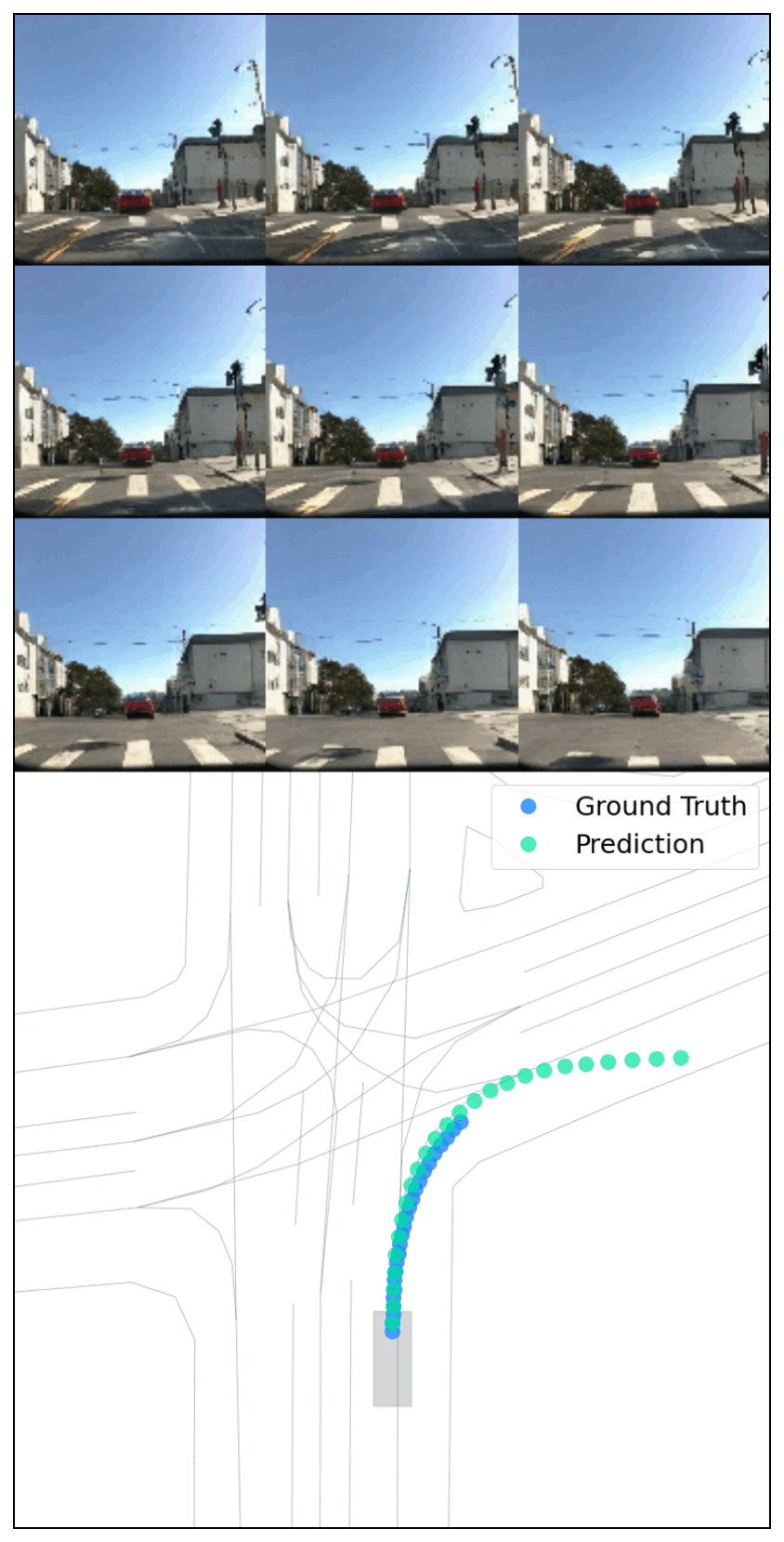}
    \begin{minipage}{.48\linewidth}\centering \footnotesize{(a) Real video as input}\end{minipage}
    \begin{minipage}{.48\linewidth}\centering \footnotesize{(b) Generated video as input}\end{minipage}
    \caption{Comparison of predicted trajectories from a planner given real and generated videos. Same scene layouts in two videos lead to highly similar trajectory predictions.}
    \label{fig:planner_traj}
  \vspace{-0.1in}

\end{figure}

Table~\ref{tab:ade-log-gen} presents the Average Displacement Error (ADE) when specific conditioning inputs are removed during video generation. Our model is trained with random dropout of conditions to promote robustness to missing inputs. The results show that removing scene layout information, specifically, bounding boxes and road maps, significantly increases ADE, highlighting the crucial role of spatial structure in guiding future trajectory predictions.

In contrast, removing operational conditions such as weather and time-of-day has a smaller impact on ADE, as long as the scene layout remains consistent. This aligns with our intuition that the surrounding geometry is the primary factor influencing ego motion. This effect is further illustrated in Fig.~\ref{fig:planner_traj}, where similar scene layouts in real and generated videos result in comparable predicted trajectories. These findings emphasize the importance of layout-aware conditioning in trajectory-aware video generation.

Table~\ref{tab:ade-change-weather} presents the planner's performance across different weather conditions. Performance slightly degrades in rainy conditions compared to no-rain conditions, as indicated by the higher ADE scores in all time horizons. This suggests that adverse weather introduces additional uncertainty that may slightly hinder accurate trajectory forecasting.

Similarly, Table~\ref{tab:ade-change-time} illustrates how the planner's performance changes across different times of day. The planner achieves its best results at noon (12:00), while performance drops slightly at midnight (00:00), likely due to reduced visibility or lighting variance in nighttime scenes. These analyses offer valuable insights into the planner's sensitivity to environmental factors, pointing to areas where targeted enhancements could strengthen model robustness under challenging conditions, as we will highlight in Sec.~\ref{sec:exp_improve}.

\begin{table}
  \centering
  \caption{ADE scores under varied weather. [\textbf{Best}, \underline{Worst}].}
  \label{tab:ade-change-weather}
  \begin{tabular}{@{}p{1.8cm}ccc@{}}
    \toprule
    Weather & ADE@1s & ADE@3s & ADE@5s \\
    \midrule
    No rain & \textbf{0.0299} & \textbf{0.2853} & \textbf{0.8580} \\
    Rain & \underline{0.0303} & \underline{0.2910} & \underline{0.8736} \\
    \bottomrule
  \end{tabular}

\end{table}

\begin{table}
  \centering
  \caption{ADE scores under varied time. [\textbf{Best}, \underline{Worst}].}
  \label{tab:ade-change-time}
  \begin{tabular}{@{}p{1.8cm}ccc@{}}
    \toprule
    Time-of-day & ADE@1s & ADE@3s & ADE@5s \\
    \midrule
    00:00 & 0.0301 & \underline{0.2907} & \underline{0.8760} \\
    06:00 & 0.0301 & \underline{0.2907} & 0.8744 \\
    12:00 & \underline{0.0302} & \textbf{0.2886} & \textbf{0.8653} \\
    18:00 & \textbf{0.0298} & 0.2893 & 0.8747 \\
    \bottomrule
  \end{tabular}
\end{table}

\subsection{Improving Planner with Generated Videos}
\label{sec:exp_improve}

We conduct experiments to evaluate the effectiveness of our generated data to fine-tune the planner. We compare two fine-tuning approaches: one is simply fine-tuning the planner on real-videos with 40K steps, and the other is to fine-tune on one million synthetic videos for 20K steps and then on real videos for 20K steps. The synthetic videos are generated with the same conditions as the real ones and the ground truth future trajectories are the same. 
We evaluate the planner's performance on real-world data to demonstrate the effectiveness of synthetic data in improving real-world performance. Table~\ref{tab:ade-ft-planner} presents the ADE of the different models. While fine-tuning solely on real data yields limited performance improvements, incorporating synthetic data from our generator effectively reduces the ADE at 5 seconds from 0.7548 to 0.7333. This demonstrates the potential of generating synthetic videos to enhance the performance of end-to-end planners.

\begin{table}
  \centering
  \caption{ADE scores on real-world validation data, fine-tuned on different data mixtures. Here, ``gen" refers to videos generated by our model.}
  \label{tab:ade-ft-planner}
  \begin{tabular}{@{}lccc@{}}
    \toprule
    Models & ADE@1s & ADE@3s & ADE@5s \\
    \midrule
    Train on real & 0.0288 & 0.2606 & 0.7548 \\
    Fine-tune on real & 0.0287 & 0.2591 & 0.7469 \\
    Fine-tune on gen + real & \textbf{0.0282} & \textbf{0.2543} & \textbf{0.7333} \\
    \bottomrule
  \end{tabular}
 
\end{table}

\begin{table}
  \centering
  \caption{ADE scores on real-world validation data with rainy weather, fine-tuned on different data mixtures.}
  \label{tab:ade-ft-planner-rain}
  \begin{tabular}{@{}lccc@{}}
    \toprule
    Models & ADE@1s & ADE@3s & ADE@5s \\
    \midrule
    Train on real & \textbf{0.0318} & 0.2893 & 0.8536 \\
    Fine-tune on real & 0.0328 & 0.2920 & 0.8482 \\
    Fine-tune on gen + real & \textbf{0.0318} & \textbf{0.2891} & \textbf{0.8382} \\
    \bottomrule
  \end{tabular}
  
\end{table}

\begin{table}
  \centering
  \caption{ADE on real-world validation data at nighttime (22:00 to 04:00), fine-tuned on different data mixtures.}
  \label{tab:ade-ft-planner-night}
  \begin{tabular}{@{}lccc@{}}
    \toprule
    Models & ADE@1s & ADE@3s & ADE@5s \\
    \midrule
    Train on real & \textbf{0.0275} & 0.2470 & 0.7372 \\
    fine-tune on real & 0.0284 & 0.2505 & 0.7328 \\
    fine-tune on gen + real & 0.0278 & \textbf{0.2447} & \textbf{0.7101} \\
    \bottomrule
  \end{tabular}
\end{table}

We further evaluate the planner’s performance in out-of-distribution scenarios, specifically rainy weather and nighttime (22:00 to 04:00). As shown in Table~\ref{tab:ade-ft-planner-rain}, fine-tuning the planner on both generated and real-world data significantly improves performance in rainy conditions compared to using real-world data alone. Similarly, Table~\ref{tab:ade-ft-planner-night} demonstrates that combining generated and real-world data for fine-tuning yields improved performance at longer time horizons (3s and 5s) for nighttime scenarios.
Notably, real-world nighttime data involves a complex interplay of factors such as traffic density and illumination, which can affect planner performance. In some cases, reduced nighttime traffic can make planning simpler, resulting in lower ADE than those in Table~\ref{tab:ade-ft-planner}. These observations highlight the challenge of isolating individual factors when relying solely on real-world data and show the advantage of our co-evaluation framework with a controllable video generation model.

\begin{figure}[t]
    \centering
    \begin{subfigure}[b]{0.236\textwidth} %
        \centering
        \begin{minipage}{0.485\textwidth}
        \centering
        \includegraphics[width=\textwidth]{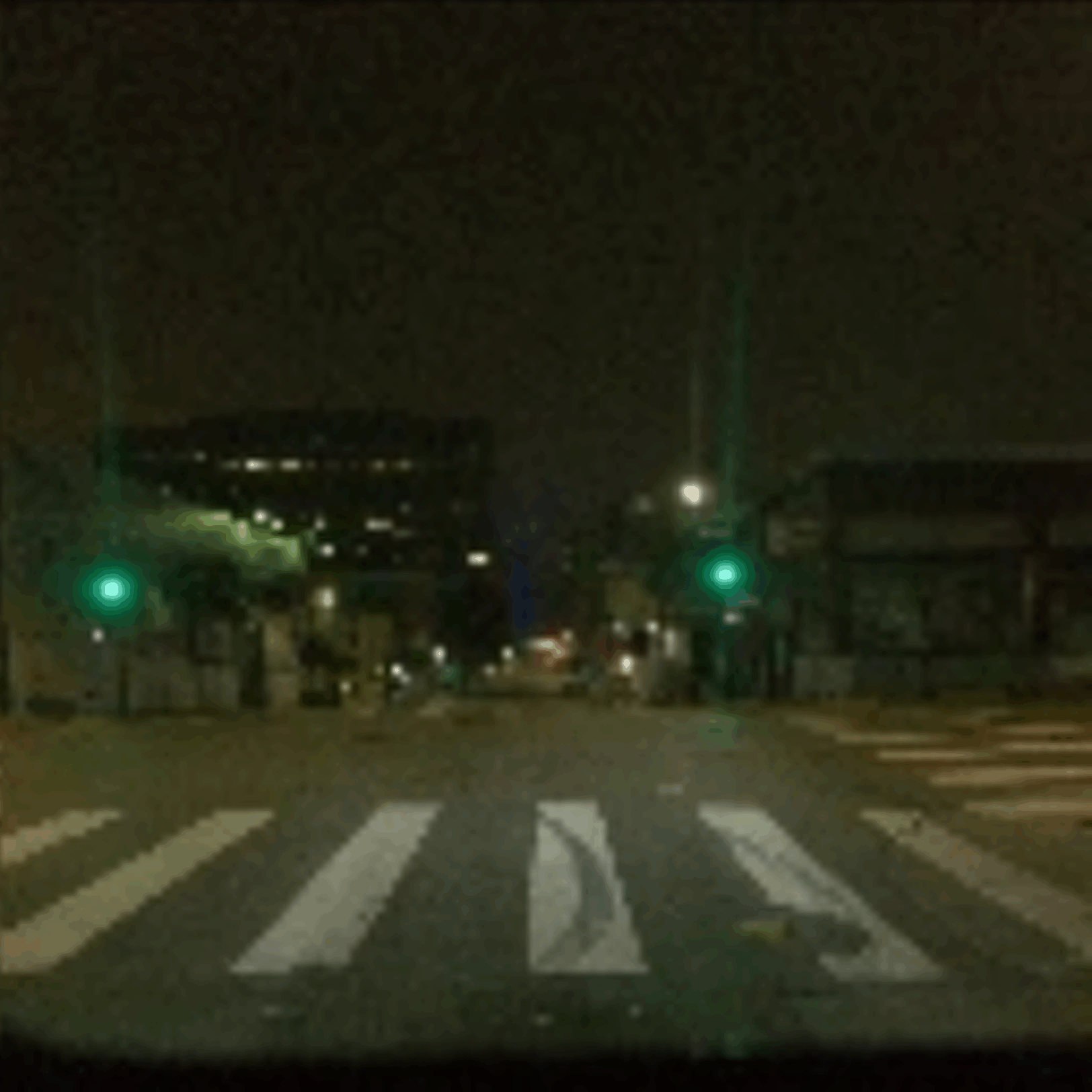}
        \caption*{Before FT}
        \end{minipage}
        \begin{minipage}{0.485\textwidth}
        \includegraphics[width=\textwidth]{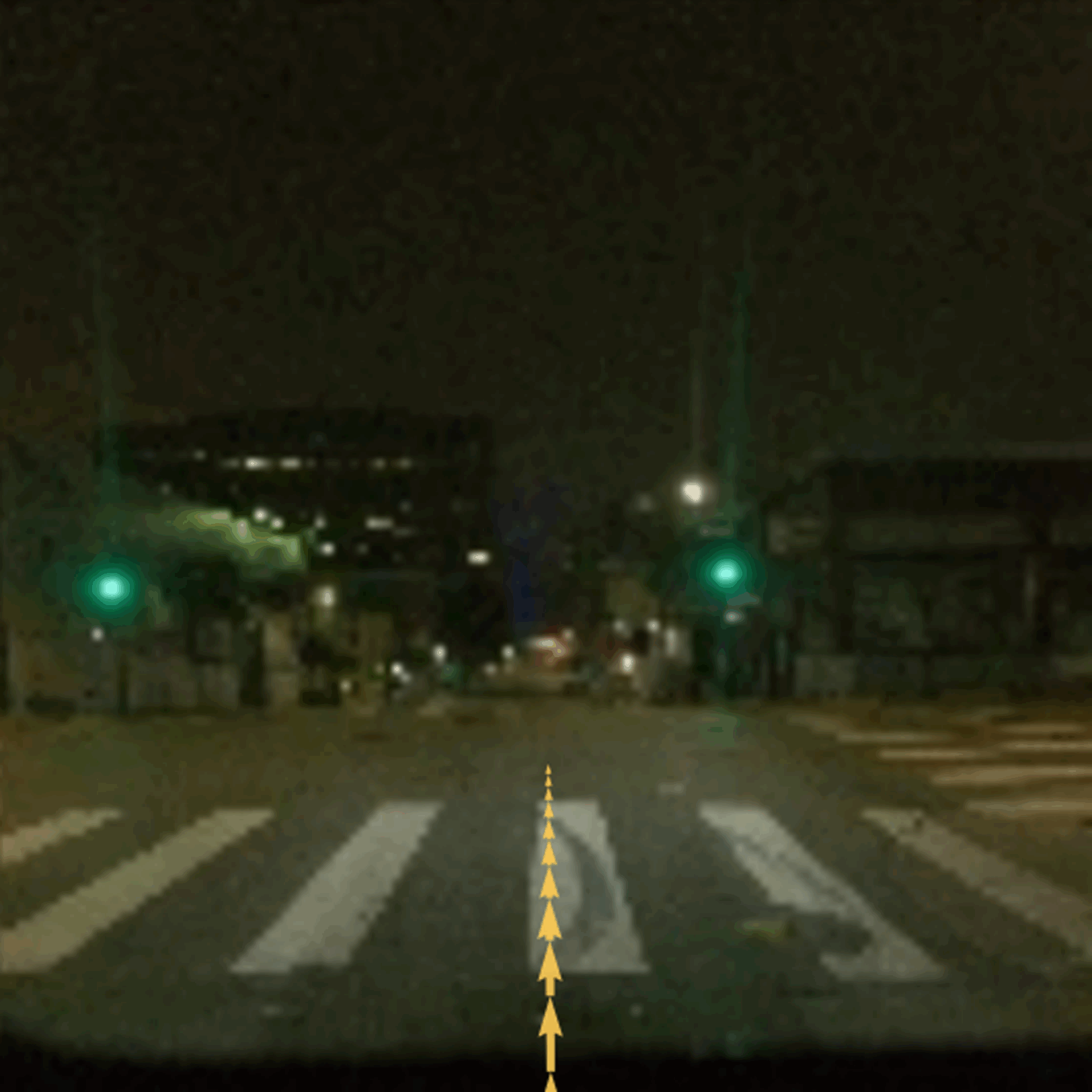}
        \caption*{After FT}
        \end{minipage}
        \caption{Case 1}
        \label{fig:image1}
        
    \end{subfigure}
    \begin{subfigure}[b]{0.236\textwidth} %
        \centering
        \begin{minipage}{0.485\textwidth}
        \centering
        \includegraphics[width=\textwidth]{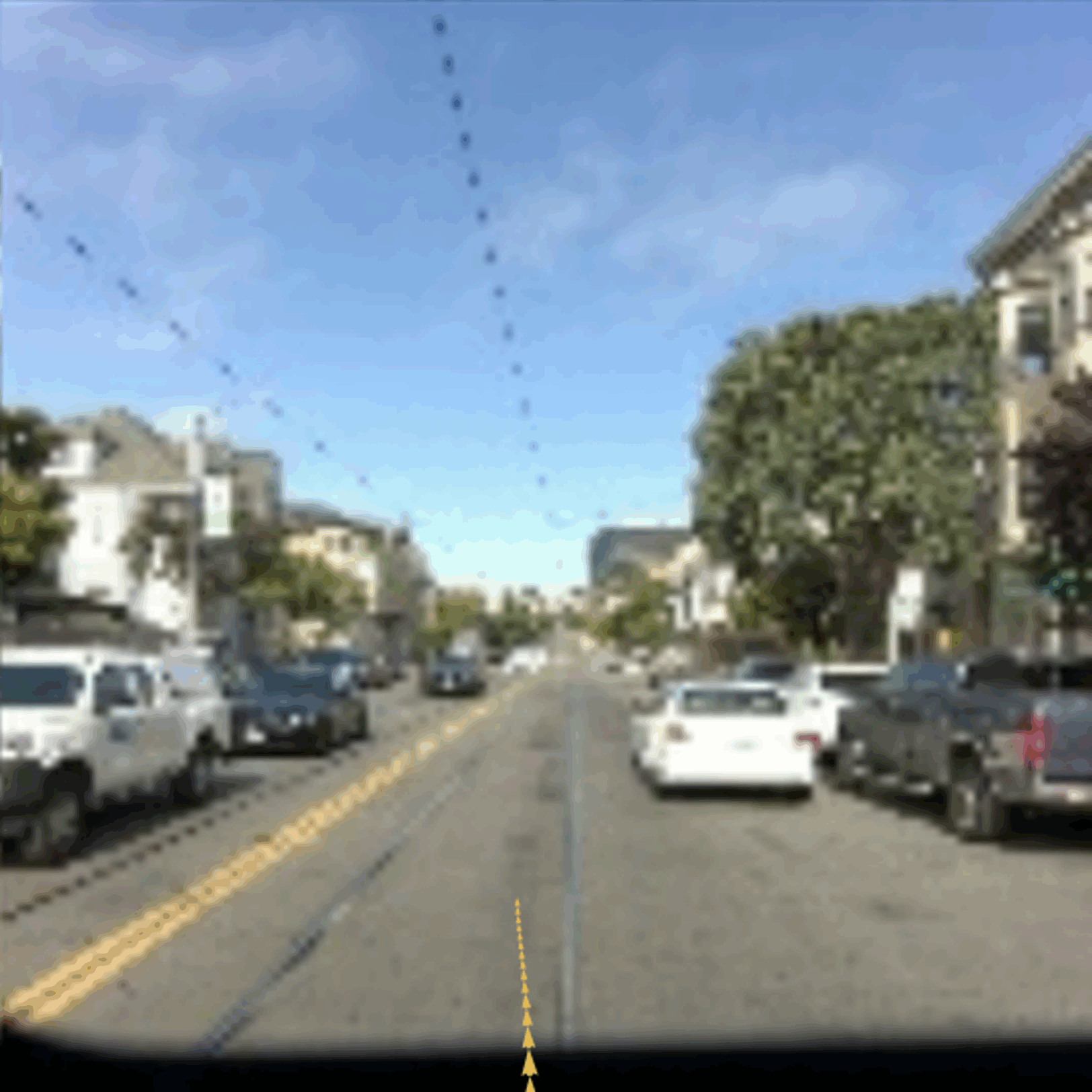}
        \caption*{Before FT}
        \end{minipage}
        \begin{minipage}{0.485\textwidth}
        \includegraphics[width=\textwidth]{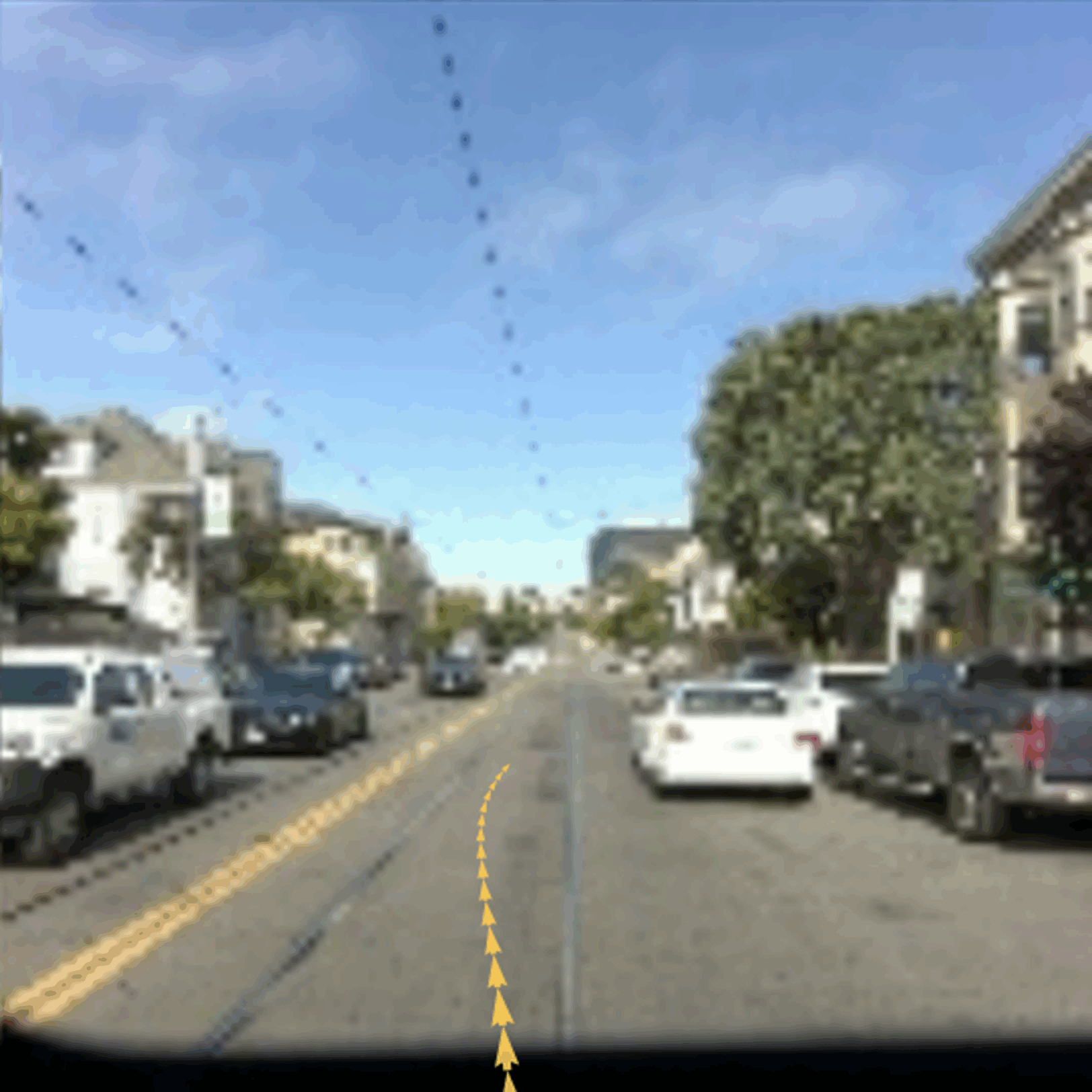}
        \caption*{After FT}
        \end{minipage}
        \caption{Case 2}
        \label{fig:image2}
    \end{subfigure}
   \caption{Qualitative results illustrating the impact of synthetic data on planner performance (yellow arrows). ``FT" means fine-tuning using synthetic and real data. Case 1: The ego vehicle's response to a green light (stopping  vs. proceeding). Case 2: The ego vehicle's interaction with a stopped vehicle in the right lane (slow movement vs. safe bypass).}

   \label{fig:rebuttal}
\end{figure}

Qualitative results in Figure~\ref{fig:rebuttal} show that fine-tuning the planner on generated and real-world data leads to improved performance in real-world driving scenarios.

\section{Limitations and Discussions}
The proposed BPT, by focusing on the distribution of planner outputs under the assumption of identical ground truth trajectories across varying environmental contexts, does not inherently assess the fidelity of physical realism or the implications for road safety. We leave a deeper investigation of these aspects to future work. While neither model is flawless and no single metric can fully capture the complexity of real-world driving, our work provides a meaningful methodology to systematically assess each component.

\section{Conclusion}

In this work, we introduce a novel framework for \textbf{co-evaluating} driving video generation and E2E planning. We propose a new metric, the Behavior Permutation Test (BPT), to assess video realism by analyzing the distribution of outputs from a planner. To the best of our knowledge, this is the \textbf{first} attempt to evaluate driving video generation using a VLM-based driving model.
In addition, we employ a video generation model with precise control over scene layout and operating conditions (e.g., weather and time of day), enabling systematic evaluation of the E2E planner. Finally, we demonstrate that synthetic data generated by our model can improve E2E planner generalization in out-of-distribution scenarios. We hope our findings will move the field closer to robustly co-evaluating and improving both generative realism and planner performance in the future.
\newpage

{
    \small
    \bibliographystyle{IEEEtran}
    \bibliography{IEEEfull}

\begin{thebibliography}{10}
\providecommand{\url}[1]{#1}
\csname url@rmstyle\endcsname
\providecommand{\newblock}{\relax}
\providecommand{\bibinfo}[2]{#2}
\providecommand\BIBentrySTDinterwordspacing{\spaceskip=0pt\relax}
\providecommand\BIBentryALTinterwordstretchfactor{4}
\providecommand\BIBentryALTinterwordspacing{\spaceskip=\fontdimen2\font plus
\BIBentryALTinterwordstretchfactor\fontdimen3\font minus \fontdimen4\font\relax}
\providecommand\BIBforeignlanguage[2]{{%
\expandafter\ifx\csname l@#1\endcsname\relax
\typeout{** WARNING: IEEEtran.bst: No hyphenation pattern has been}%
\typeout{** loaded for the language `#1'. Using the pattern for}%
\typeout{** the default language instead.}%
\else
\language=\csname l@#1\endcsname
\fi
#2}}

\bibitem{hwang2024emmaendtoendmultimodalmodel}
J.-J. Hwang, R.~Xu, H.~Lin, W.-C. Hung, J.~Ji, K.~Choi, D.~Huang, T.~He, P.~Covington, B.~Sapp, \emph{et~al.}, ``Emma: End-to-end multimodal model for autonomous driving,'' \emph{arXiv preprint arXiv:2410.23262}, 2024.

\bibitem{tian2024drivevlm}
X.~Tian, J.~Gu, B.~Li, Y.~Liu, Y.~Wang, Z.~Zhao, K.~Zhan, P.~Jia, X.~Lang, and H.~Zhao, ``Drivevlm: The convergence of autonomous driving and large vision-language models,'' \emph{arXiv preprint arXiv:2402.12289}, 2024.

\bibitem{ho2022video}
J.~Ho, T.~Salimans, A.~Gritsenko, W.~Chan, M.~Norouzi, and D.~J. Fleet, ``Video diffusion models,'' in \emph{NeurIPS}, 2022.

\bibitem{videoworldsimulators2024}
T.~Brooks, B.~Peebles, C.~Holmes, W.~DePue, Y.~Guo, L.~Jing, D.~Schnurr, J.~Taylor, T.~Luhman, E.~Luhman, \emph{et~al.}, ``Video generation models as world simulators,'' \emph{OpenAI Blog}, vol.~1, p.~8, 2024.

\bibitem{gupta2023photorealisticvideogenerationdiffusion}
A.~Gupta, L.~Yu, K.~Sohn, X.~Gu, M.~Hahn, F.-F. Li, I.~Essa, L.~Jiang, and J.~Lezama, ``Photorealistic video generation with diffusion models,'' in \emph{ECCV}, 2024.

\bibitem{goodfellow2014explaining}
I.~J. Goodfellow, J.~Shlens, and C.~Szegedy, ``Explaining and harnessing adversarial examples,'' \emph{arXiv preprint arXiv:1412.6572}, 2014.

\bibitem{Montali23neurips_wosac}
N.~Montali, J.~Lambert, P.~Mougin, A.~Kuefler, N.~Rhinehart, M.~Li, C.~Gulino, T.~Emrich, Z.~Yang, S.~Whiteson, B.~White, and D.~Anguelov, ``The waymo open sim agents challenge,'' in \emph{NeurIPS}, 2023.

\bibitem{LeCun2022world}
Y.~LeCun, ``A path towards autonomous machine intelligence version 0.9. 2, 2022-06-27,'' 2022.

\bibitem{NEURIPS2018_2de5d166}
D.~Ha and J.~Schmidhuber, ``Recurrent world models facilitate policy evolution,'' in \emph{NeurIPS}, 2018.

\bibitem{hafner2019dreamer}
D.~Hafner, T.~Lillicrap, J.~Ba, and M.~Norouzi, ``Dream to control: Learning behaviors by latent imagination,'' \emph{arXiv preprint arXiv:1912.01603}, 2019.

\bibitem{hafner2021mastering}
D.~Hafner, T.~P. Lillicrap, M.~Norouzi, and J.~Ba, ``Mastering atari with discrete world models,'' in \emph{ICLR}, 2021.

\bibitem{hafner2023dreamerv3}
D.~Hafner, J.~Pasukonis, J.~Ba, and T.~Lillicrap, ``Mastering diverse domains through world models,'' \emph{arXiv preprint arXiv:2301.04104}, 2023.

\bibitem{Kim2020_GameGan}
S.~W. Kim, Y.~Zhou, J.~Philion, A.~Torralba, and S.~Fidler, ``{Learning to Simulate Dynamic Environments with GameGAN},'' in \emph{CVPR}, 2020.

\bibitem{singer2023makeavideo}
U.~Singer, A.~Polyak, T.~Hayes, X.~Yin, J.~An, S.~Zhang, Q.~Hu, H.~Yang, O.~Ashual, O.~Gafni, D.~Parikh, S.~Gupta, and Y.~Taigman, ``Make-a-video: Text-to-video generation without text-video data,'' in \emph{ICLR}, 2023.

\bibitem{wu2023tune}
J.~Z. Wu, Y.~Ge, X.~Wang, S.~W. Lei, Y.~Gu, Y.~Shi, W.~Hsu, Y.~Shan, X.~Qie, and M.~Z. Shou, ``Tune-a-video: One-shot tuning of image diffusion models for text-to-video generation,'' in \emph{ICCV}, 2023.

\bibitem{blattmann2023videoldm}
A.~Blattmann, R.~Rombach, H.~Ling, T.~Dockhorn, S.~W. Kim, S.~Fidler, and K.~Kreis, ``Align your latents: High-resolution video synthesis with latent diffusion models,'' in \emph{CVPR}, 2023.

\bibitem{blattmann2023stablevideodiffusionscaling}
A.~Blattmann, T.~Dockhorn, S.~Kulal, D.~Mendelevitch, M.~Kilian, D.~Lorenz, Y.~Levi, Z.~English, V.~Voleti, A.~Letts, \emph{et~al.}, ``Stable video diffusion: Scaling latent video diffusion models to large datasets,'' \emph{arXiv preprint arXiv:2311.15127}, 2023.

\bibitem{Peebles2022DiT}
W.~Peebles and S.~Xie, ``Scalable diffusion models with transformers,'' in \emph{ICCV}, 2023.

\bibitem{hu2023gaia1generativeworldmodel}
A.~Hu, L.~Russell, H.~Yeo, Z.~Murez, G.~Fedoseev, A.~Kendall, J.~Shotton, and G.~Corrado, ``Gaia-1: A generative world model for autonomous driving,'' \emph{arXiv preprint arXiv:2309.17080}, 2023.

\bibitem{ma2024unleashinggeneralizationendtoendautonomous}
E.~Ma, L.~Zhou, T.~Tang, Z.~Zhang, D.~Han, J.~Jiang, K.~Zhan, P.~Jia, X.~Lang, H.~Sun, \emph{et~al.}, ``Unleashing generalization of end-to-end autonomous driving with controllable long video generation,'' \emph{arXiv preprint arXiv:2406.01349}, 2024.

\bibitem{wang2023drivedreamerrealworlddrivenworldmodels}
X.~Wang, Z.~Zhu, G.~Huang, X.~Chen, J.~Zhu, and J.~Lu, ``Drivedreamer: Towards real-world-drive world models for autonomous driving,'' in \emph{ECCV}, 2024.

\bibitem{wen2024panacea}
Y.~Wen, Y.~Zhao, Y.~Liu, F.~Jia, Y.~Wang, C.~Luo, C.~Zhang, T.~Wang, X.~Sun, and X.~Zhang, ``Panacea: Panoramic and controllable video generation for autonomous driving,'' in \emph{CVPR}, 2024.

\bibitem{wang2024driving}
Y.~Wang, J.~He, L.~Fan, H.~Li, Y.~Chen, and Z.~Zhang, ``Driving into the future: Multiview visual forecasting and planning with world model for autonomous driving,'' in \emph{CVPR}, 2024.

\bibitem{gao2024vista}
S.~Gao, J.~Yang, L.~Chen, K.~Chitta, Y.~Qiu, A.~Geiger, J.~Zhang, and H.~Li, ``Vista: A generalizable driving world model with high fidelity and versatile controllability,'' in \emph{NeurIPS}, 2024.

\bibitem{li2023drivingdiffusion}
X.~Li, Y.~Zhang, and X.~Ye, ``Drivingdiffusion: Layout-guided multi-view driving scenarios video generation with latent diffusion model,'' in \emph{ECCV}, 2024.

\bibitem{yang2024genad}
J.~Yang, S.~Gao, Y.~Qiu, L.~Chen, T.~Li, B.~Dai, K.~Chitta, P.~Wu, J.~Zeng, P.~Luo, J.~Zhang, A.~Geiger, Y.~Qiao, and H.~Li, ``{Generalized Predictive Model for Autonomous Driving},'' in \emph{CVPR}, 2024.

\bibitem{gao2023magicdrive}
R.~Gao, K.~Chen, E.~Xie, L.~Hong, Z.~Li, D.-Y. Yeung, and Q.~Xu, ``Magicdrive: Street view generation with diverse 3d geometry control,'' in \emph{ICLR}, 2024.

\bibitem{nuscenes}
H.~Caesar, V.~Bankiti, A.~H. Lang, S.~Vora, V.~E. Liong, Q.~Xu, A.~Krishnan, Y.~Pan, G.~Baldan, and O.~Beijbom, ``nuscenes: A multimodal dataset for autonomous driving,'' in \emph{CVPR}, 2020.

\bibitem{Mumuni_2024}
\BIBentryALTinterwordspacing
A.~Mumuni, F.~Mumuni, and N.~K. Gerrar, ``A survey of synthetic data augmentation methods in machine vision,'' \emph{Machine Intelligence Research}, vol.~21, no.~5, p. 831–869, Mar. 2024. [Online]. Available: \url{http://dx.doi.org/10.1007/s11633-022-1411-7}
\BIBentrySTDinterwordspacing

\bibitem{bowles2018gan}
C.~Bowles, L.~Chen, R.~Guerrero, P.~Bentley, R.~Gunn, A.~Hammers, D.~A. Dickie, M.~V. Hern{\'a}ndez, J.~Wardlaw, and D.~Rueckert, ``Gan augmentation: Augmenting training data using generative adversarial networks,'' \emph{arXiv preprint arXiv:1810.10863}, 2018.

\bibitem{ma2023generating}
W.~Ma, Q.~Liu, J.~Wang, A.~Wang, X.~Yuan, Y.~Zhang, Z.~Xiao, G.~Zhang, B.~Lu, R.~Duan, \emph{et~al.}, ``Generating images with 3d annotations using diffusion models,'' in \emph{ICLR}, 2024.

\bibitem{xu2017end}
H.~Xu, Y.~Gao, F.~Yu, and T.~Darrell, ``End-to-end learning of driving models from large-scale video datasets,'' in \emph{CVPR}, 2017.

\bibitem{hu2022st}
S.~Hu, L.~Chen, P.~Wu, H.~Li, J.~Yan, and D.~Tao, ``St-p3: End-to-end vision-based autonomous driving via spatial-temporal feature learning,'' in \emph{ECCV}, 2022.

\bibitem{hu2023planning}
Y.~Hu, J.~Yang, L.~Chen, K.~Li, C.~Sima, X.~Zhu, S.~Chai, S.~Du, T.~Lin, W.~Wang, \emph{et~al.}, ``Planning-oriented autonomous driving,'' in \emph{CVPR}, 2023.

\bibitem{casas2021mp3}
S.~Casas, A.~Sadat, and R.~Urtasun, ``Mp3: A unified model to map, perceive, predict and plan,'' in \emph{CVPR}, 2021.

\bibitem{gu2023vip3d}
J.~Gu, C.~Hu, T.~Zhang, X.~Chen, Y.~Wang, Y.~Wang, and H.~Zhao, ``Vip3d: End-to-end visual trajectory prediction via 3d agent queries,'' in \emph{CVPR}, 2023.

\bibitem{jiang2023vad}
B.~Jiang, S.~Chen, Q.~Xu, B.~Liao, J.~Chen, H.~Zhou, Q.~Zhang, W.~Liu, C.~Huang, and X.~Wang, ``Vad: Vectorized scene representation for efficient autonomous driving,'' in \emph{ICCV}, 2023.

\bibitem{sadat2020perceive}
A.~Sadat, S.~Casas, M.~Ren, X.~Wu, P.~Dhawan, and R.~Urtasun, ``Perceive, predict, and plan: Safe motion planning through interpretable semantic representations,'' in \emph{ECCV}, 2020.

\bibitem{wang2024omnidrive}
S.~Wang, Z.~Yu, X.~Jiang, S.~Lan, M.~Shi, N.~Chang, J.~Kautz, Y.~Li, and J.~M. Alvarez, ``Omnidrive: A holistic llm-agent framework for autonomous driving with 3d perception, reasoning and planning,'' \emph{arXiv preprint arXiv:2405.01533}, 2024.

\bibitem{bai20243dtokenizedllmkeyreliable}
Y.~Bai, D.~Wu, Y.~Liu, F.~Jia, W.~Mao, Z.~Zhang, Y.~Zhao, J.~Shen, X.~Wei, T.~Wang, \emph{et~al.}, ``Is a 3d-tokenized llm the key to reliable autonomous driving?'' \emph{arXiv preprint arXiv:2405.18361}, 2024.

\bibitem{wei2022chainofthoughtpromptingelicitsreasoning}
J.~Wei, X.~Wang, D.~Schuurmans, M.~Bosma, F.~Xia, E.~Chi, Q.~V. Le, D.~Zhou, \emph{et~al.}, ``Chain-of-thought prompting elicits reasoning in large language models,'' \emph{NeurIPS}, 2022.

\bibitem{chen2023pali}
X.~Chen, J.~Djolonga, P.~Padlewski, B.~Mustafa, S.~Changpinyo, J.~Wu, C.~R. Ruiz, S.~Goodman, X.~Wang, Y.~Tay, \emph{et~al.}, ``Pali-x: On scaling up a multilingual vision and language model,'' \emph{arXiv preprint arXiv:2305.18565}, 2023.

\bibitem{chen2024end}
L.~Chen, P.~Wu, K.~Chitta, B.~Jaeger, A.~Geiger, and H.~Li, ``End-to-end autonomous driving: Challenges and frontiers,'' \emph{IEEE Transactions on Pattern Analysis and Machine Intelligence}, 2024.

\bibitem{jia2024bench2drive}
X.~Jia, Z.~Yang, Q.~Li, Z.~Zhang, and J.~Yan, ``Bench2drive: Towards multi-ability benchmarking of closed-loop end-to-end autonomous driving,'' \emph{arXiv preprint arXiv:2406.03877}, 2024.

\bibitem{stylegan_v}
I.~Skorokhodov, S.~Tulyakov, and M.~Elhoseiny, ``Stylegan-v: A continuous video generator with the price, image quality and perks of stylegan2,'' in \emph{CVPR}, 2022.

\bibitem{unterthiner2018towards}
T.~Unterthiner, S.~Van~Steenkiste, K.~Kurach, R.~Marinier, M.~Michalski, and S.~Gelly, ``Towards accurate generative models of video: A new metric \& challenges,'' \emph{arXiv preprint arXiv:1812.01717}, 2018.

\bibitem{nayakanti2023wayformer}
N.~Nayakanti, R.~Al-Rfou, A.~Zhou, K.~Goel, K.~S. Refaat, and B.~Sapp, ``Wayformer: Motion forecasting via simple \& efficient attention networks,'' in \emph{ICRA}, 2023.

\bibitem{jaegle2021perceiver}
A.~Jaegle, S.~Borgeaud, J.-B. Alayrac, C.~Doersch, C.~Ionescu, D.~Ding, S.~Koppula, D.~Zoran, A.~Brock, E.~Shelhamer, \emph{et~al.}, ``Perceiver io: A general architecture for structured inputs \& outputs,'' in \emph{ICLR}, 2022.

\bibitem{shi2023zero123++}
R.~Shi, H.~Chen, Z.~Zhang, M.~Liu, C.~Xu, X.~Wei, L.~Chen, C.~Zeng, and H.~Su, ``Zero123++: a single image to consistent multi-view diffusion base model,'' \emph{arXiv preprint arXiv:2310.15110}, 2023.

\end{thebibliography}
}

\end{document}